\documentclass[12pt]{article}
\usepackage[utf8]{inputenc}
\usepackage{adjustbox}
\usepackage{url}
\usepackage{hyperref} 
 
\usepackage{amsthm,bm,amsfonts,amsmath}
\usepackage{graphicx,psfrag,epsf}
\usepackage{enumerate}
\usepackage[authoryear]{natbib}
\usepackage{xr}
\newtheorem{thm}{Theorem}

\newtheorem{Corollary}{Corollary}

\newtheorem{Example}{Example}

\newcommand\remove{\bgroup\markoverwith{\textcolor{gray}{\rule[.5ex]{2pt}{1pt}}}\ULon}

\newcommand{\x}{\mathbf{x}}
\newcommand{\aug}{\text{aug}}
\renewcommand{\a}{\text{a}}
\newcommand{\X}{\mathbf{X}}

\renewcommand{\P}{\mathbb{P}}
\newcommand{\I}{\mathbb{I}}
\newcommand{\E}{\mathbb{E}}

\def\X{{\mathbf{X}}}
\def\x{{\mathbf{x}}}

\def\E{{\mathbb E}}

\begin{document}
\def\spacingset#1{\renewcommand{\baselinestretch}%
{#1}\small\normalsize} \spacingset{1}

{
  \title{\bf Is augmentation effective to improve prediction in imbalanced text datasets?}
  \author{Gabriel O. Assunção\\
    Department of Statistics, Universidade Federal de Minas Gerais, \\ Belo Horizonte, Brazil\\
    and\\
    Rafael Izbicki \\
    Department of Statistics, Universidade Federal de São Carlos, \\ São Carlos, Brazil \\
    and \\
    Marcos O. Prates\\
    Department of Statistics,  Federal de Minas Gerais, \\Belo Horizonte, Brazil
    }
  \maketitle
} 
\bigskip

\begin{abstract}
 Imbalanced datasets present a significant challenge for machine learning models, often leading to biased predictions. To address this issue, data augmentation techniques are widely used in natural language processing (NLP) to generate new samples for the minority class. However, in this paper, we challenge the common assumption that data augmentation is always necessary to improve predictions on imbalanced datasets. Instead, we argue that adjusting the classifier cutoffs without data augmentation can produce similar results to oversampling techniques. Our study provides theoretical and empirical evidence to support this claim.  Our findings contribute to a better understanding of the strengths and limitations of different approaches to dealing with imbalanced data, and help researchers and practitioners make informed decisions about which methods to use for a given task.
\end{abstract}

\noindent%
\vfill

\newpage
\spacingset{1.45}
\section{Introduction}

Imbalanced datasets are a widespread issue encountered in real-life text datasets, where certain classes contain a higher number of observations than others. For instance, positive reviews of a product tend to outweigh negative ones. However, machine learning models trained on imbalanced data can lead to biased predictions, and addressing this concern is crucial. This is particularly pertinent in natural language processing (NLP), where imbalanced classes arise in applications such as spam filtering \citep{al2014isrd} and sentiment analysis \citep{wang2013sample}. Similarly, in image classification, detecting objects in images with minimal examples poses a significant challenge \citep{gao2014enhanced}. 

Dealing with imbalanced data can be achieved through various approaches, which can be classified into three categories: preprocessing methods, cost-sensitive methods, and algorithmic methods \citep{kaur2019systematic}. In NLP, one of the most commonly used approaches is preprocessing, which involves generating new samples to balance the dataset. This method is also known as sampling, and it aims to generate new data synthetically for the minority class.

Two popular sampling methods are Synthetic Minority Oversampling Technique (SMOTE) \citep{chawla2002smote} and Random Oversampling, which have been extensively used and studied \citep{li2010data, padurariu2019dealing}. These techniques increase the representation of minority classes, which can reduce bias towards the majority class and improve the model's performance \citep{feng2021survey}. 
Indeed, such techniques are recommended by hundreds of blog posts  and papers \citep{brownlee_2021,Mustafa_2019,Yogesh_2019,abdoh2018cervical, tan2019wireless, rupapara2021impact, akkaradamrongrat2019text, mohasseb2018improving, tesfahun2013intrusion}.
Other  approaches include the Easy Data Augmentation (EDA) technique \citep{wei2019eda}, which applies simple functions to generate new text in the minority class.
 
Oversampling techniques are widely used to enhance prediction accuracy, especially when the classification rule relies on assigning a new text to the label with the highest estimated probability \citep{abdoh2018cervical, tan2019wireless, rupapara2021impact, akkaradamrongrat2019text, mohasseb2018improving, tesfahun2013intrusion}. However, 
in this paper, we argue that contrary to common belief, data augmentation is usually unnecessary to improve predictions  on imbalanced datasets. Our argument is that oversampling techniques are widely used because  
 most machine learning softwares
 rely on assigning a new instance to the label with the highest estimated probability by default. This is not optimal.
Indeed, we show that by changing how the  probabilities estimated using the
unaugmented data are used to create a classifier,  we can maximize the classification performance for most datasets.   As a result, the purported benefits of data augmentation may be misleading.

To support our argument, we provide both theoretical and empirical evidence that challenges the common belief that data augmentation is always beneficial for imbalanced datasets. We believe that our findings will contribute to a better understanding of the strengths and limitations of different approaches to dealing with imbalanced data, and will help researchers and practitioners to make more informed decisions about which methods to use for a given task.

Let $Y$ denote the label of interest, and $\x$ is the vector of features. For simplicity, let us assume that $Y$ takes binary values, $0$ or $1$. In this study, we prove that if we can accurately estimate $\P(Y=1|\x)$, changing the cutoffs of the classifier without data augmentation produces the same results as using Random Oversampling (Section \ref{sec:fund}). That is, classifying an instance as belonging to the positive class based on $\P(Y=1|\x)>c$, where $c$ is properly chosen (see Section \ref{sec:fund} for details), is equivalent to performing data augmentation and classifying with a cutoff of $0.5$, the default majority rule used by most software.

\begin{figure}[!h]
\centering
    \includegraphics[width=.9\textwidth]{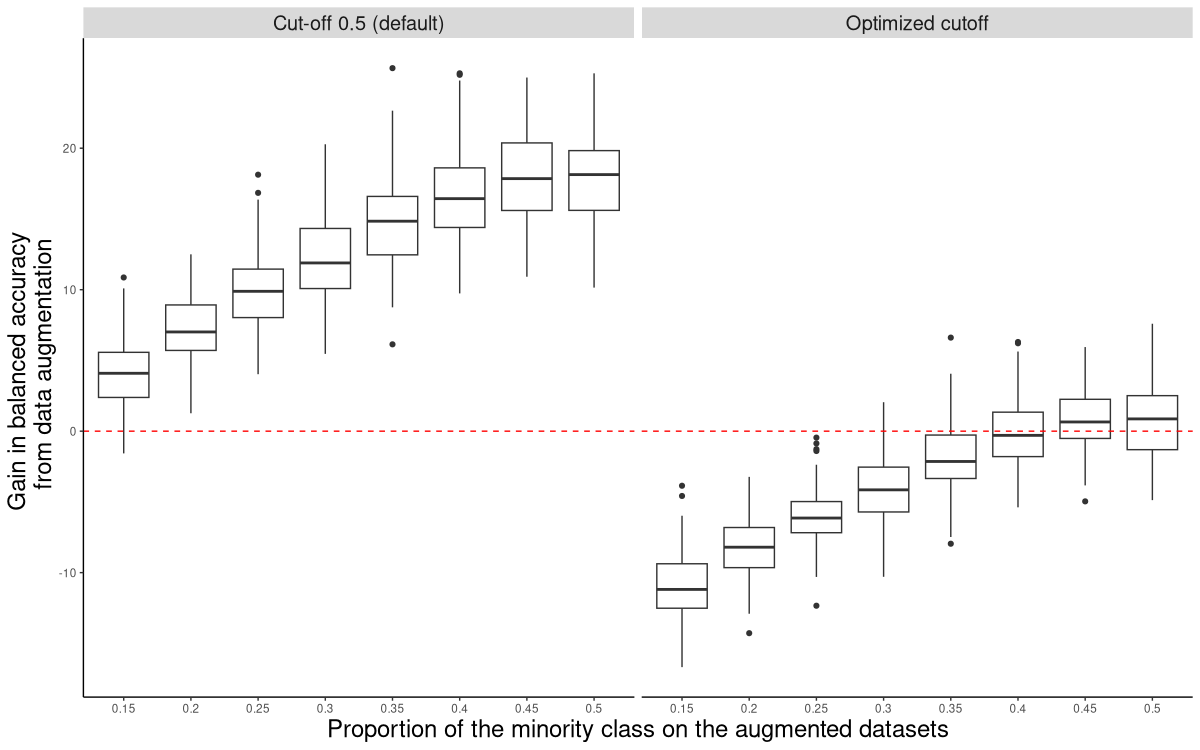}
    \caption{Gain in balanced accuracy when performing data augmentation for the Amazon Review Dataset (see Section \ref{sec:exp} for details). If the default cutoff $c=0.5$ for $\widehat P(Y=1|\x)$ is chosen to perform classification, it appears that data augmentation is effective (left panel). However, the right panel reveals that when $c$ is appropriately chosen, there is no improvement in performance through data augmentation}
    \label{fig:intro_result}
\end{figure}

To illustrate this point, Figure \ref{fig:intro_result} presents the results obtained from the Amazon Review dataset \citep{keung2020multilingual}. The left panel of the figure shows the improvement in balanced accuracy achieved by a model that uses Random Oversampling compared to a base model that does not use augmentation when a threshold of $0.5$ is used for classification. However, as shown in the right panel, by properly choosing the value of $c$, we can obtain similar results without using data augmentation.

Therefore, the only way  random oversampling
can produce better classifications is if it can create a more accurate estimate of $\P(y|\x)$. However, there is  no evidence to support the fact that $\P(y|\x)$ will be better estimated using oversampling: oversampling only reuses the same training points, so there is no new information on the augmented dataset. Indeed, Section \ref{sec:exp} presents empirical data that suggests random oversampling does not improve the accuracy of the probability estimate. Moreover, that section also goes beyond  Random Oversampling  and provides empirical evidence
that even other state-of-the-art augmentation methods  achieve similar results to a model without oversampling as long as the cutoff is properly chosen. Finally, Section~\ref{s:con} ends with a conclusion and discussion of our discoveries.

\section{Random Oversampling}
\label{sec:fund}

Let $\X \in \mathcal{X}$ denote the features (covariates), and $Y \in \{0, ..., K-1 \} := \mathcal{Y}$ be the label of interest. We start by proving that the probabilities $\P(Y=k|\x)$ (henceforth the \emph{conditional probabilities}) induced by the Random Oversampling technique have a one-to-one relationship with the original probabilities associated with the non-augmented dataset. We will show that, in the binary case, this implies that any classification rule based on the augmented conditional probabilities (such as classifying an instance as positive if its conditional probability is larger than 0.5) is equivalent to a rule based on the non-augmented probabilities; one only needs to select an appropriate threshold point. We will assume we know $\P$. That is, it does not need to be estimated. 
All proofs are shown in the Supplementary Material S.1.

Let $\P$ be the probability distribution associated with the original (non-augmented) data $$\mathcal{T} = \{ (\X_1, Y_1), ..., (\X_n, Y_n) \}.$$

We denote the dataset with augmented samples by $$\mathcal{T}' = \{ (\X'_1, Y'_1), ..., (\X'_m, Y'_m) \}.$$

We assume each sample from $\mathcal{T}'$ is sampled using the following rule:
\begin{enumerate}
    \item Draw $k \in \mathcal{Y}$, such that each element of $\mathcal{Y}$ has probability $w_k$ of being drawn, where $w_k$'s are predefined non-negative numbers such that $\sum_{k=0}^{K-1} w_k=1$.
    \item Randomly choose an observation from $\{(\X_i, Y_i) \in \mathcal{T}: Y_i=k\}$.
\end{enumerate}

The augmented dataset is then given by merging $\mathcal{T}$ with $\mathcal{T}'$: $$\mathcal{T}_{\aug} = \mathcal{T} \cup \mathcal{T}'.$$
This scheme is a generalization of vanilla random oversampling:

\begin{Example}[Random Oversampling for binary labels]
\label{ex:binary}
In a binary classification problem $\mathcal{Y} = \{0, 1\}$, suppose that label 1 is the minority class. To oversample the data, we set $w_1 = 1$ and $w_0 = 0$, and to make the augmented data to be 50-50, we choose $m=n(\P(Y=0)-\P(Y=1))$.
\end{Example}

This data-generating scheme, together with $\P$, induce a distribution over each $(\X, Y) \in \mathcal{T}_{\aug}$. The following theorem shows how this distribution, denoted by $\P_{\a}$, relates to $\P$.

\begin{thm}
\label{thm:relationship}
For each $j \in \mathcal{Y}$,
\begin{align}
\label{eq:p_a}
   \P_{\a}(Y=j|\x)=\frac{\frac{\P_{\a}(Y=j)}{\P(Y=j)}\P(Y=j|\x)}{\sum_{k=0}^{K-1} \frac{\P_{\a}(Y=k)}{\P(Y=k)}\P(Y=k|\x)}, 
\end{align}
where
$$\P_{\a}(Y=j)=\frac{n}{n+m}\P(Y=j)+ \frac{m}{n+m}w_j.$$

Conversely, it holds that
\begin{align*}
   \P(Y=j|\x)=\frac{\frac{\P(Y=j)}{\P_{\a}(Y=j)}\P_{\a}(Y=j|\x)}{\sum_{k=1}^K \frac{\P(Y=k)}{\P_{\a}(Y=k)}\P_{\a}(Y=k|\x)}. 
\end{align*}
\end{thm}

In practice, classification rules for probabilistic classifiers are typically obtained by minimizing the expected value of a  loss function $L: \mathcal{Y} \times  \mathcal{Y} \longrightarrow \mathbb{R}$, where  $L(k, j)$ represents the loss of classifying an instance of the class k as being of the class $j$ and  $L(k, k) = 0$ for every $k \in \mathcal{Y}$. The optimal classifier under $\P_\a$ is given by \citep{ripley2007pattern}
\begin{align}
    \label{eq:optimal_Pa}
    g^*(\x):= \arg \min_{j \in \mathcal{Y}} \sum_{k \in \mathcal{Y}}  L(k, j) \P_\a(Y=k|\x).
\end{align}

In the binary case, the optimal decision rule corresponds to checking whether the probability is greater than the threshold $\frac{L(0, 1)}{L(0, 1)+L(1, 0)}$ as presented in Theorem~\ref{thm:binary}.

\begin{thm}[\citep{ripley2007pattern}; page 19]
\label{thm:binary}
In the binary case (that is, $K=2$), criteria \ref{eq:optimal_Pa} leads to the following decision rule:
\begin{equation}
\label{eq:binary_optimal}
    g^*(\x)=
    \begin{cases}
      1, & \text{if}\  \P_\a(Y=1|\x)\geq \frac{L(0, 1)}{L(0, 1) + L(1, 0)} \\
      0, & \text{otherwise}.
    \end{cases}
  \end{equation}
\end{thm}

In practice, the decision rule used is often to choose the label that maximizes the $\P_a(Y=k|\x)$. This classifier is a special case when the errors have the same cost, this classifier is presented in Theorem \ref{thm:zeroone}.

\begin{thm}[\citep{ripley2007pattern}; page 19]
\label{thm:zeroone}
Under the 0-1 loss (that is, all misclassification errors have the same cost $L(k,j)=\I(k \neq j)$), criteria \ref{eq:optimal_Pa} leads to the following optimal classifier: $$g^*(\x):= \arg \min_{j \in \mathcal{Y}} \sum_{k \in \mathcal{Y}| k \neq j} \P_\a(Y=k|\x),$$ which, in the binary case, is equivalent to 
\begin{equation}
\label{eq:optimal_binary}
    g^*(\x)=
    \begin{cases}
      1, & \text{if}\  \P_\a(Y=1|\x)\geq \frac{1}{2} \\
      0, & \text{otherwise}.
    \end{cases}
  \end{equation}
\end{thm}

The following theorem shows that we can obtain the same classifier from Equation \eqref{eq:optimal_Pa}, $g^*$, using the probability associated with the original data but with a different loss.

\begin{thm}
\label{thm:equivalence}
Let $g^*$ be the optimal classifier according to the loss $L$ and the distribution induced by the augmentation procedure $\P_\a$ (Equation \eqref{eq:optimal_Pa}). Then
$$g^*(\x)= \arg \min_{j \in \mathcal{Y}} \sum_{k \in \mathcal{Y}}  L'(k,j) \P(Y=k|\x),$$
where 
$L'(k,j)=L(k,j)\frac{\P_{\a}(Y=k)}{\P(Y=k)}$.
\end{thm}

In particular, it follows from Theorems \ref{thm:zeroone} and \ref{thm:equivalence}  that, in the binary case, the optimal classifier for the augmented dataset according to the 0-1 loss 
can be computed by comparing the original conditional probabilities 
to the threshold $\P(Y=1)$:
\begin{Corollary}
\label{cor:relation}
The classifier of Equation \eqref{eq:optimal_binary} is equivalent to
  \begin{equation}
  \label{eq:originial_prob}
    g^*(\x)=
    \begin{cases}
      1, & \text{if}\ \P(Y=1|\x) \geq  \P(Y=1)  \\
      0, & \text{otherwise}
    \end{cases}
  \end{equation}
    This classifier is the one that maximizes the balanced accuracy.
\end{Corollary}

\begin{figure}[!h]
    \centering
    \includegraphics[width=.5\textwidth]{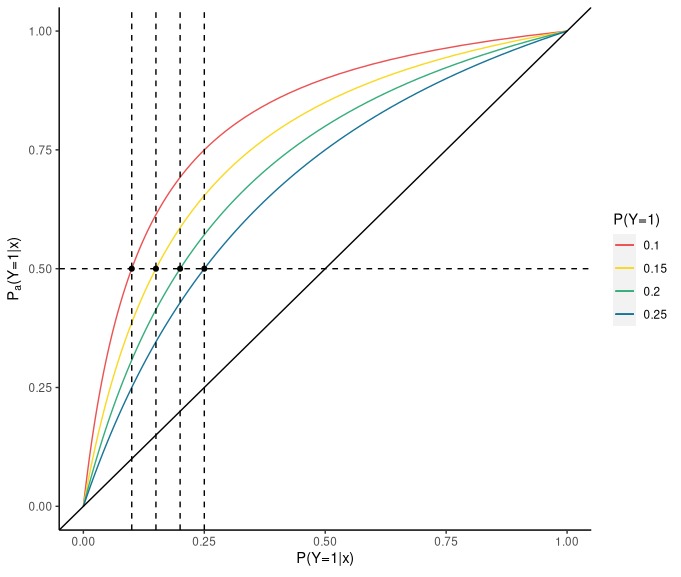}
    \caption{Relationship between  the original probability $\P(Y=1|\x)$ and the augmented probability $\P_a(Y=1|\x)$ (Theorem \ref{thm:relationship}) for different values of $\P(Y=1)$.  The dashed lines illustrate  Corollary \ref{cor:relation}: setting the threshold  on the augmented data  to 0.5 corresponds to using the threshold of $\P(Y=1)$ on the original probability.}
    \label{fig:prob}
\end{figure}

Figure \ref{fig:prob} illustrates the relationship between the original probability $\P(Y=1|\x)$ and the augmented probability $\P_a(Y=1|\x)$
for various values of $\P(Y=1)$. In particular, it illustrates Corollary \ref{cor:relation} by showing  that  $\P_a(Y=1|\x) = 0.5$ always corresponds to $\P(Y=1|\x)=\P(Y=1)$, and therefore checking whether $\P_a(Y=1|\x) > 0.5$ is equivalent to checking whether $\P(Y=1|\x)>\P(Y=1)$. 
%To demonstrate the effects of Theorem \ref{thm:equivalence} and Corollary \ref{cor:relation}, the points in the figure represents the pair $(x,y)$ when the threshold is of $0.5$ in the augmented dataset for different imbalanced percentages, e.g., when the $P(Y=1) = 0.25$ the pair $(x,y)$ is given by $(0.25,0.50)$.

Corollary \ref{cor:relation} proves that,  if we know either $\P(Y=1|\x)$ or $\P_{\a}(Y=1|\x)$, then comparing the former to an optimized threshold will yield the same loss as using a threshold of 0.5 and applying Random Oversampling.  However, in practice, we must estimate these probabilities using models, which could potentially lead to different results. Nevertheless, empirical results in the following section indicate that selecting the appropriate threshold for $\widehat{\P}(Y=1|\x)$ is sufficient to achieve the same outcome as using $\widehat \P_{\a}(Y=1|\x)$, even if the probabilities are estimated. 
This is because Random Oversampling does not provide any additional information about the dataset; it merely reuses the same sample points from the original training set. Surprisingly,  the empirical results in the next section  suggest  that this is also the case even when using more advanced and complex data augmentation methods.

\section{Experiments}
\label{sec:exp}
% Minha escrita: Next, we empirically compare several oversampling techniques on real datasets. Our goal is to show that by properly choosing the threshold, there is typically no gain in performing oversampling. The illusion of gain often comes from the fact that these thresholds are chosen as their default value. 

We perform an empirical comparison of several oversampling techniques on real datasets with the aim of demonstrating that, by appropriately selecting the threshold, there is generally no advantage in employing oversampling. The perceived benefits are often illusory and arise from the adoption of default threshold values.

In our experiments, we  use eight datasets, all available online:
\begin{itemize}
    \item Sentiment Analysis in Twitter \citep{barbieri2020tweeteval}. 
    \item Women's E-commerce Clothing Reviews \citep{agarap2018statistical}.
    \item Android Apps and User Feedback \citep{grano2017android}.
    \item Yelp reviews \citep{zhang2015character}.
    \item Multilingual Amazon review corpus \citep{keung2020multilingual}.
    \item Hate Speech Offensive \citep{davidson2017automated}
    \item Jigsaw toxicity dataset \citep{manerba2022bias}. We used two versions of this dataset using the Insult and Toxicity labels. 
\end{itemize}

Table S.1 in the Supplementary Material S.2 shows a brief description of the dataset features, minority class ratio, and the main goal of the used datasets.
%\footnote{\url{https://huggingface.co/datasets/tweet_eval}} \footnote{\url{https://www.kaggle.com/datasets/nicapotato/womens-ecommerce-clothing-reviews}} \footnote{\url{https://huggingface.co/datasets/app_reviews}} \footnote{\url{https://huggingface.co/datasets/yelp_review_full}} \footnote{\url{https://huggingface.co/datasets/amazon_reviews_multi}} \footnote{\url{https://huggingface.co/datasets/hate_speech_offensive}} \footnote{\url{https://www.kaggle.com/c/jigsaw-unintended-bias-in-toxicity-classification}}. Table \ref{tab:dataset} briefly describes the dataset.

To assess the efficacy of different oversampling methodologies, we conducted a random sampling of the original datasets five times, with two sample sizes: 500 and 2000. Each sample was utilized to generate a distinct model, and no examples were duplicated between samples. To validate the models, we established a separate validation dataset by randomly sampling from the original dataset, comprising two distinct sizes: 125  for models trained with 500 sample points and 500  for models trained with 2000  sample points. All models were compared using an independent sample with 1000  sample points.

The following oversampling methods  were used in the experiments: 
\begin{itemize}
    \item \textbf{Random Sampling}: sentences of the training dataset are randomly sampled.
    \item \textbf{Random Oversampling Examples} (ROSE) \citep{menardi2014training}. This is a smoothed version of Random Oversampling. It is controlled by a shrinkage parameter that controls the distribution to generate the new data. When this parameter is equal to zero, the method is equivalent to Random Sampling. We used the shrinkage values of 0.5, 1, and 3.
    \item \textbf{SMOTE} \citep{chawla2002smote}. This method creates synthetic data considering the neighbors of a minority class observation, the synthetic data is created between the observation and the neighbor.
    \item \textbf{Borderline SMOTE} \citep{han2005borderline}. This is a variation of SMOTE  that uses borderline samples to generate synthetic data, that is, the observation used to create the new samples are those that are misclassified by a KNN classifier \citep{fix1952discriminatory}.
    \item \textbf{Easy Data Augmentation} (EDA) \citep{wei2019eda}. This is a method that applies simple operations to the text to generate new sentences. To control the number of procedures to be applied is used a parameter $\alpha \in [0,1]$ and the length of the sentence $L$. In this article, we will use three types of EDA, the \textbf{Synonyms Replace (SR)}; \textbf{Random Insertion (RI) }; \textbf{Random Deletion (RD)}. 
    %The procedures used in this article are:
    %\begin{itemize}
    %    \item \textbf{Synonyms Replace} (SR): replacing words with synonyms. The number of words to be changed is $\alpha \times L$
    %    \item \textbf{Random Insertion} (RI): this technique inserts random words in random positions. The number of words inserted is given by $\alpha \times L$
    %    \item \textbf{Random Deletion} (RD): this technique deletes words randomly in the sentence. The probability of a word being removed is $\alpha$
    %\end{itemize}
    \item We propose a technique for generating new sentences based on the distribution of the words. We coined it \textbf{ImpOrtance Word Augmentation} (IOWA). This method is fully described in the Supplementary Material S.3. We investigate  four variants:
    \begin{itemize}
        \item \textbf{Frequency}: where we attribute more importance to the most frequent words.
        \item \textbf{The difference in frequencies}: we attribute more importance to a word that is more frequent in a label than the others.
        \item \textbf{Inverse Document-Frequency} (IDF): we attribute more importance to rare behavior words.
        \item \textbf{The difference in IDF}:  we attribute more importance to words that are rare in one group, but common in the others.
    \end{itemize}
\end{itemize}

All oversampling methods are compared using 
a random forest classifier \citep{breiman2001random} with a  Bag-of-Words input. %The Random Sampling and IOWA methods are also trained using a logistic regression on the Bag-of-Words.

To evaluate the effectiveness of a classifier $g(\x) \in {0,1}$, we use the balanced accuracy metric, which calculates the average of the sensitivity and specificity of the classifier:  $$\text{ba}(g)=\frac{\P(g(\X)=1|Y=1)+\P(g(\X)=0|Y=0)}{2}.$$ In the supplementary material we demonstrate the results for F1-score, sensitivity, specificity and accuracy.

In practice, we estimate  $\text{ba}(g)$ using the test data. To enhance interpretability, we compare the performance of the classifier obtained via data augmentation, $g_{\text{aug}}$, with that of a non-augmented classifier $g_{\text{base}}$ by reporting  the percentage gain in balanced accuracy achieved by the augmented classifier over the non-augmented one:
%As we have different datasets and in each, we use five samples to train the models the balanced accuracy of the models trained with/without the augmented technique will be in a different scale, to ignore this effect on our results we evaluate the percentage gain (Equation \eqref{eq:percentage_gain}) of using the augmentation technique over not using it,
%We used the balanced accuracy score, the mean between specificity and sensibility, to compare our methods as the primary metric.  Since each train sample and model gives us a different balanced accuracy, we evaluate the percentage gain (Equation \eqref{eq:percentage_gain}) of using the augmentation technique over not using it.
%\begin{equation}
%    \label{eq:percentage_gain}
%    PercentageGain = \frac{Balanced Accuracy_{augmented} - Balanced Accuracy_{base}}{Balanced Accuracy_{base}}.
%\end{equation}
\begin{equation}
    \label{eq:percentage_gain}
    \text{PercentageGain} = \frac{\text{ba}(g_{\text{aug}}) - \text{ba}(g_{\text{base}})}{\text{ba}(g_{\text{base}})}.
\end{equation}

%We increase the minority class of all train datasets to get a 50-50 ratio 40 times, with each technique. To check whether the augmented method in fact increases the quality of the model, we perform a hypothesis test to test whether  the percentage gain is  zero (see [APENDICE/MATERIAL SUPLEMENTAR] for details about this hypothesis test). 

We also test whether the augmentation methods improve the estimates of the probabilities $\P(Y=1|\x)$ by using the percentage gain (Equation \eqref{eq:percentage_gain}) in terms of the Area Under ROC  (AUC) \citep{lusted1971decision} and the Brier Score \citep{brier1950verification}:
\begin{equation*}
    \label{eq:brier_score}
    \text{BrierScore} = \E\left[(Y - \hat{\P}(Y=1|\X))^2\right].
\end{equation*}
Again we use the test data to estimate such quantities. Moreover, to make results meaningful, to compute such scores  we first map the augmented estimates $\widehat \P_a(Y=1|\x)$ back to 
$\widehat \P(Y=1|\x)$ using the last equation of Theorem \ref{thm:relationship}.

We augmented all train datasets 40 times using each technique to achieve a 50-50 ratio for the minority class. To assess whether the augmented method truly enhances the model's performance, we conducted a hypothesis test to ascertain whether the percentage gain is zero (for further information regarding this hypothesis test, refer to the Supplementary Material S.4).

%To illustrate the results of Section \ref{sec:fund} we will demonstrate what happens based on the threshold choice, first, we used the well-known threshold choice of 0.5 to compare the methods; second, we use the result of Corollary \ref{cor:relation} to compare the theoretical threshold on the base model that is similar to the threshold of 0.5 on the augmented model; after that, we will demonstrate what happens when we choose a threshold with the same rule on the augmented method. 

%We compare three
%In order to explicate the outcomes of Section \ref{sec:fund}, we shall exemplify the implications of varying threshold values. Specifically, we will commence by implementing the established threshold of 0.5 to evaluate the methodologies. Subsequently, we will employ the findings of Corollary \ref{cor:relation} to contrast the theoretical threshold of the underlying model with that of the augmented model, equivalent to the threshold of 0.5. Following this, we will demonstrate the repercussions of selecting a threshold in congruence with the aforementioned rule for the augmented method.

%To evaluate if the methods in fact increase the probability estimation we will use the percentage gain (Equation \eqref{eq:percentage_gain}) with the ROC Area Under Curve (AUC) and Brier Score (Equation \eqref{eq:brier_score}) [Adicioar referencia].

%At least we will give an overall analysis of the methods to illustrate if the model is better using the AUC metric and the Brier Score.

\subsection{The importance of choosing a suitable threshold}

To evaluate the significance of selecting an appropriate threshold, we employed two classification rules in our methodology:
\begin{itemize}
    \item \textbf{(c=0.5)} The first rule is  the widely-used classifier that  assigns labels based on the class with a higher probability. Because we only deal with binary classification, this corresponds to using  the value of $c=0.5$ as a threshold. 
    \item  \textbf{(Optimized c)}  The second rule, inspired by Corollary \ref{cor:relation}, involves choosing a threshold $c$ that maximizes the balanced accuracy on a validation dataset. 

\end{itemize}

\begin{figure}[!h]
\centering
    \includegraphics[width=.9\textwidth]{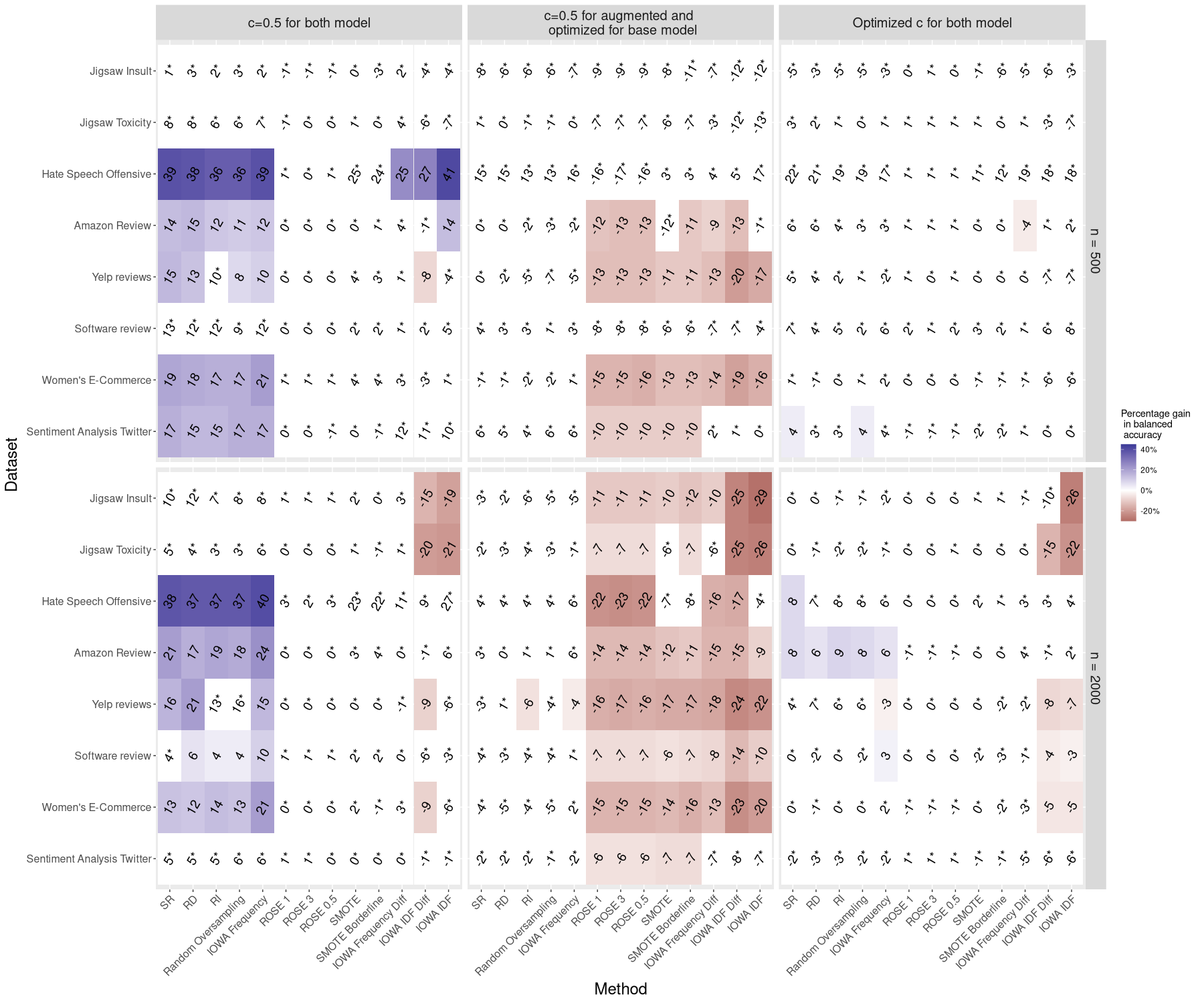}
    \caption{Heatmap of the mean percentage gain in balanced accuracy when comparing the augmented methods with the non-augmented model for classification rules.  Positive values indicate superior performance by the augmented method. Non-significant, at a significance level of 1\%, gains are marked with asterisks and displayed in white. Our findings indicate that data augmentation provides a noticeable benefit only when using the default threshold of $c=0.5$ (left column); optimizing the threshold on non-augmented data eliminates the need for augmentation.}
    \label{fig:heatmap}
\end{figure}

We perform three comparative evaluations of the following classifiers:
\begin{enumerate}
    \item \textbf{(c=0.5 for augmented and base models)} We evaluate the standard classifier  that uses $c=0.5$ on both augmented and non-augmented models. This is the approach usually taken by blog posts and papers that advocate for the use of data augmentation, as well as the default options in most machine learning packages, such as Python Scikit-learn models \citep{scikit-learn}; XGBoost API \citep{rxgboost}; R Random Forest package \citep{rrandom}
    \item \textbf{(c=0.5 for the augmented model, and optimized for base model)} We compare the non-augmented method, but now using the threshold that optimizes the balanced accuracy,  against the augmented method with a classification rule of 0.5. The goal is to check whether choosing $c$ in the non-augmented model is enough to achieve the same accuracy as we would get by doing data augmentation. This is what we expect to happen on  the Random Oversampling method (Corollary \ref{cor:relation}). 
    \item \textbf{(Optimized c augmented and base models)} We optimize the cutoffs for both the base and the augmented models.
    This comparison aims to determine whether the other techniques have any advantages over the non-augmented model that could not be achieved by choosing an appropriate threshold.   \end{enumerate} 
Figure \ref{fig:heatmap} displays the average percentage gain attained for the augmented techniques in each dataset. The top rows represent the results achieved on models trained with a sample size of 500, while the bottom rows correspond to models trained with a sample size of 2000. Results demonstrating no statistical significance are identified by an asterisk and presented against a white background. Significance is denoted by a color scale, wherein red denotes instances where the non-augmented method produces superior outcomes and blue indicates cases in which the augmented method yields better results.

%As expected,
 %in the setting where $c=0.5$ for both classifiers (first column), we observe that
 %EDA, IOWA using frequency, and Random Oversampling techniques seem to significantly improve the non-augmented classifier both for $n=500$ and $n=2,000$. 
 %, in both sample sizes. It is noteworthy that the augmented data often generates higher Balanced Accuracy values on average, which may lead to the assumption that the classifiers, after being augmented, are better at discriminating between labels. This is a common misconception that is frequently utilized as evidence in favor of applying these methods.
%However,
%Corollary \ref{cor:relation} establishes the theoretical result that, in the case of Random Oversampling, the same classifier can be obtained by selecting an appropriate cut-off point on the non-augmented model. 
%the second column  shows that this is not the case if the cutoff is optimized for the base model. Indeed, essentially all data-augmentation techniques yielded either non-significant improvements in the balanced accuracy, and some even made it worse.
%, or even 
%Notably, the Random Oversampling technique did not yield a significant percentage gain in this comparison. 
%Furthermore, the techniques that performed better on the augmented model using the default rule did not achieve a significant gain, and the technique that showed no substantial improvement even with the default now yielded a lower balanced accuracy, significantly so for the models trained with a sample size of 2000.

In the first column, where both classifiers have c = 0.5, we observe that EDA, IOWA using frequency, and Random Oversampling techniques trained on the augmented data are significantly better than the base classifier,  both for n = 500 and n = 2,000, on most datasets. However, when the cutoff is optimized for the base model, the second column shows that essentially all data augmentation techniques either yield non-significant improvements in balanced accuracy or even make it worse. Thus, the apparent benefit of data augmentation in the first column is illusory.

Furthermore, we optimized the threshold on the augmented sample to see if data augmentation could benefit the classifiers in the third column. However, the results indicate that this did not help either. In all but one case, selecting a good cutoff for the base model was sufficient, and data augmentation did not improve the performance of the classifiers.

In Section S.5 of the Supplementary Material the same analysis for a logistic model is presented for the Random Oversampling and IOWA methods, for the logistic regression even when using the threshold of 0.5 the augmentation technique did not demonstrate improvement.

\subsection{Does data-augmentation improve the estimate of $\P(Y=y|\x)$?}
The preceding findings demonstrate the influence of data augmentation on classification accuracy through a comparison of $\widehat \P(Y=1|\x)$ against a threshold. We conclude that data augmentation scarcely enhances performance if the threshold is properly selected on a non-augmented model. It is important to note, however, that a lack of improvement in the metrics does not necessarily imply a lack of improvement in the estimation. This section aims to explore the potential of data augmentation in improving the estimate of $\P(Y=1|\x)$.

\begin{figure}[!h]
    \centering
    \includegraphics[width=.9\textwidth]{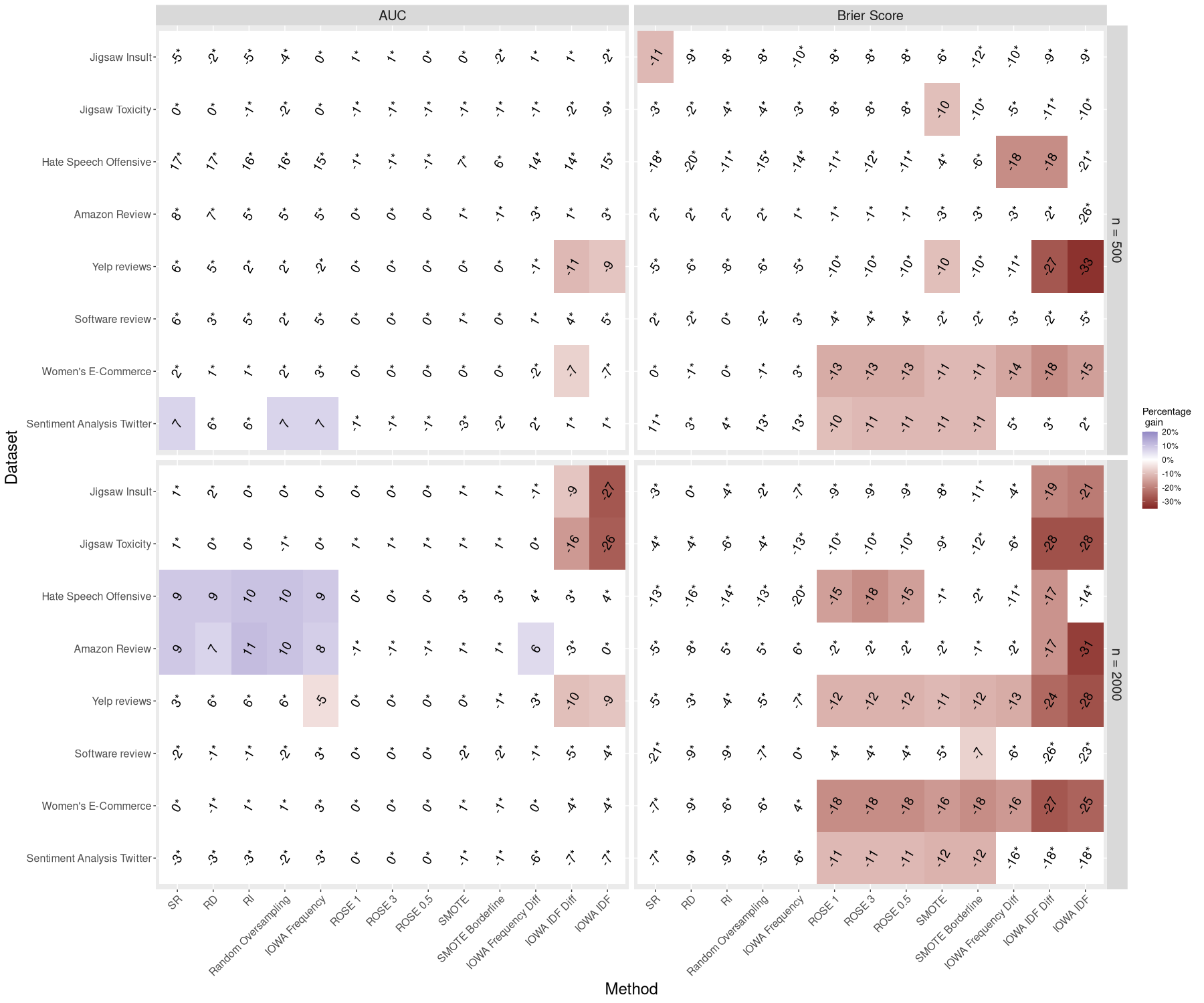}
    \caption{Heatmap of the average percentage improvement in the AUC (left column) and Brier Score (right column) when comparing the augmented methods with the non-augmented ones.  Positive values indicate superior precision in estimating $\P(Y=1|\x)$ using the augmented method. Non-significant, at a significance level of 1\%, gains are marked with asterisks and displayed in white. The AUC results indicate that data augmentation only rarely improves $\P(Y=1|\x)$ estimates, while the Brier Score suggests that it never improves them and often leads to worse results.}
    \label{fig:heatmap_metrics}
\end{figure}

The left column of Figure \ref{fig:heatmap_metrics} shows the percentage gain in AUC for the augmented models. As before, results are marked with asterisks to indicate non-significant percentage gains, while values with a blue background signify significant improvements and those with a red background indicate a significant lack of improvement.
Our results reveal that, in almost all cases, the gain in AUC for augmented models was non-significant. This indicates that there is often little to no improvement in estimating $\P(Y=1|\x)$ when using data augmentation. 
%In particular, there was no gain in the AUC metric when $n=2,000$.

Next, we compare the behavior of ROC curves for the base and the augmented models. As there are 40 trials per train repetition,  we use functional boxplots \citep{functionalboxplot}. The results for the Yelp dataset with a sample size of 500 are illustrated in Figure \ref{fig:roc_curve}, for the others dataset see Supplementary Material S.6. 
The plots show that almost all ROC curves
 for the non-augmented model are within the purple region, indicating that, in most cases, it does not differ from the ROC curves of the augmented models. The only two exceptions to these are for the methods IOWA using IDF and IOWA using Difference IDF, in which the  non-augmented models have slightly better ROC curves.
 
%In cases where the AUC percentage gain is positive, indicating an enhancement in estimation, the median of the ROC curves in the augmented model surpasses that of the ROC curve for the non-augmented model. As the percentage gain decreases, the curves tend to decrease as well, and the median ROC curve of the augmented model approaches that of the non-augmented model. Conversely, when the gain is negative, the curve of the non-augmented model exceeds the median curve of the augmented model, indicating a lack of improvement in the estimation.

\begin{figure}[!h]
    \centering
    \includegraphics[width=.9\textwidth]{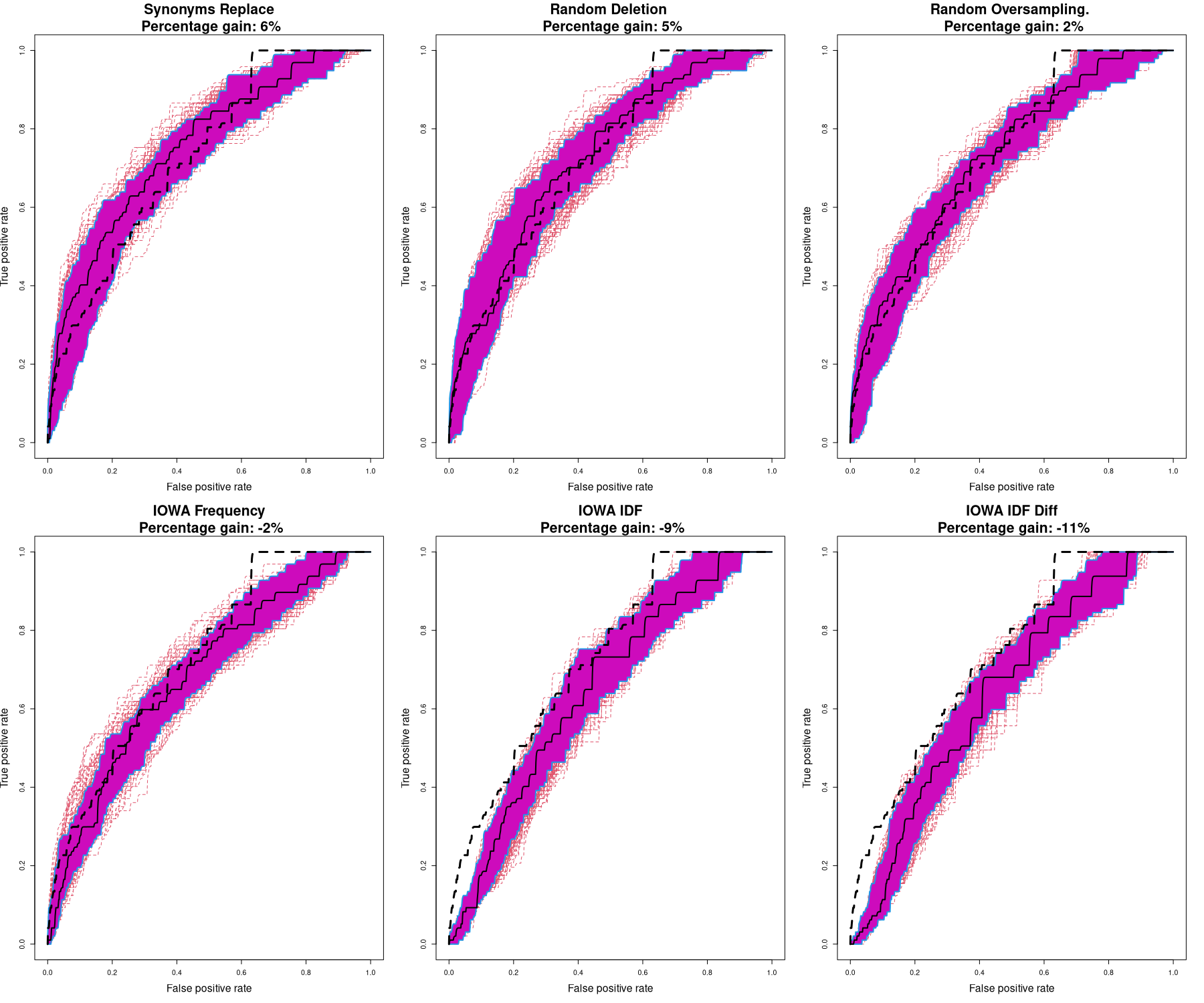}
    \caption{Functional Boxplot of the ROC curves for the Yelp Dataset on the train size of 500. The blue curves, the straight black curve, and the red curve define, respectively, the interval, median, and outliers of the ROC curve for the augmented model, the purple area bounded by the blue curves represents the inter-quartile range; the dashed black curve represents the mean ROC curve of the base model. There is no benefit in doing data augmentation.}
    \label{fig:roc_curve}
\end{figure}

%When dealing with imbalanced data, a more appropriate metric to evaluate a model is the Brier Score (Equation \ref{eq:brier_score}). Since the augmented methods generate probabilities in the augmented space, we utilized Theorem \ref{thm:relationship} to recover the original probability and compared the Brier Scores of the methods to assess if the estimations improve. 
%The bottom row of Figure \ref{fig:heatmap_metrics} displays the percentage gain in the Brier Score for the augmented model.

%When dealing with imbalanced data a better metric to be used to evaluate a model is the Brier Score (Equation \ref{eq:brier_score}). Since the augmented methods generate the probability in the augmented space, we used Theorem \ref{thm:relationship} to retrieve the original probability and compare the methods' Brier Score to analyze if the estimations increase. 
We next evaluated the performance of estimating $\P(Y=1|\x)$ using the Brier Score (right column of Figure \ref{fig:heatmap_metrics}). Our results suggest that data augmentation did not lead to a statistically significant improvement in estimating $\P(Y=1|\x)$. In fact, in many cases, it resulted in worse performance.

%many cases, the non-augmented models have a significantly greater Brier Score, suggesting that there is no improvement in the estimation of $\P(Y=1|\x)$ using an augmentation technique. Although the EDA method and the IOWA using frequency show similar probability estimation results, a slight improvement is observed in the case of small datasets; however, the improvement is not deemed statistically significant.

%Those results show evidence that the models combined with the augmented methods do not generate a better estimation of the probabilities. In fact, we have the perception of the improvement based on the classification rule, but when we analyze the probabilities estimation does not have an improvement. 

\section{Conclusions}
\label{s:con}
By utilizing both theoretical derivations and empirical evaluations,   we  conclude that selecting a threshold that optimizes specific metrics of the model directly in the imbalanced dataset is enough to achieve good prediction accuracy; there is no benefit in doing data augmentation. This conclusion is reinforced by the comparable results achieved between the balanced accuracy metric and the F1-Score, as detailed in Supplementary Material S.7. Furthermore, our comprehensive analysis indicates that incorporating augmentation methods into the model does not yield a superior estimation of probabilities. 

%For the IOWA using Frequency and EDA in some cases, we achieved greater results optimizing the threshold on the augmented model, but this doesn't happen often. This could be justified by the fact that the augmentation is helping improve the class probability estimation, however, this is scarcely observed and further investigation on what are the data characteristics that this phenomenon happens must be investigated. In this paper, we did not observe the impact of the $\alpha$ parameter in the EDA method for oversampling, in future works this parameter should be considered.  The other methods, in general, do not achieve superior results.  

The results of our research are aligned with those obtained by \citet{van2022harm}, who conducted a simulated investigation to examine the impact of widely-used imbalanced correction techniques in the context of predicting ovarian cancer diagnosis. These results demonstrate that beyond natural language processing, the perceived benefits of utilizing augmented techniques may also be misleading. 

Although  we found that none of the data augmentation methods were able to substantially improve the performance of estimates of $\P(y|x)$ and resulting classifiers, this does not mean that it is impossible to develop methods that can improve performance. In fact, if it were possible to generate new data in an i.i.d. fashion, estimates of $\P(y|x)$ (and therefore the performance of resulting classifiers) would be superior. This raises the question of how to perform data augmentation in a way that approximates i.i.d. and opens up the discussion of new techniques to improve performance.
%For instance, 
% a promising augmentation strategy that has not been examined in this study and could be applied in natural language processing involves employing language models for text generation, such as those proposed by \citet{kobayashi2018contextual,sennrich2015improving,yang2020generative}. 

%In future work, we can extend for other methods a more general relation of the probabilities on the augmented and the original sets to seek a way to use these methods to achieve better results. For example, find another way to define the optimum threshold or how to generate samples to improve the estimation of the class probability systematically. Additionally, a promising augmentation strategy that has not been examined in this study and could be applied in natural language processing involves employing language models for text generation, such as those proposed in \cite{kobayashi2018contextual}, \cite{sennrich2015improving}, and \cite{yang2020generative}. 

\subsection*{Acknowledgements} 
Marcos O. Prates would like to acknowledge (Conselho Nacional de Desenvolvimento Científico e Tecnológico) CNPq grants 436948/2018-4 and 307547/2018-4 and FAPEMIG (Fundação de Amparo à Pesquisa do Estado de Minas Gerais) grant APQ-01837-22 and CAPES (Coordenação de Aperfeiçoamento de Pessoal de Nível Superior) for financial support. 
Rafael Izbicki is grateful for the financial support of CNPq (309607/2020-5 and 422705/2021-7) and FAPESP (2019/11321-9).

\bibliographystyle{apalike}
\bibliography{refs}

\end{document}

\def\spacingset#1{\renewcommand{\baselinestretch}%
{#1}\small\normalsize} \spacingset{1}

{
  \title{\bf Supplementary material to the paper ``Is augmentation effective to improve prediction in imbalanced text datasets?"}
  \author{Gabriel O. Assunção\\
    Department of Statistics, Universidade Federal de Minas Gerais, \\ Belo Horizonte, Brazil\\
    and\\
    Rafael Izbicki \\
    Department of Statistics, Universidade Federal de São Carlos, \\ São Carlos, Brazil\\
    and \\
    Marcos O. Prates\\
    Department of Statistics,  Federal de Minas Gerais, \\Belo Horizonte, Brazil
    }
  \maketitle
}

\section{Proofs}
\label{sec:sup_th}
In this section, we will present the proof of Theorems and Corollary of Section 2. 

\begin{proof}[Proof of Theorem 1]
\label{eq:proof_a}

We assume $x$ is discrete, although the proof is always valid.
$$\P_{\mathcal{T}}(X=x|Y=k)=\P_{\mathcal{T'}}(X=x|Y=k).$$
It follows that
\begin{equation*}
    \begin{split}
        \P_{a}(\{Y=k\} \cap \{X=x\}) &= \frac{n}{n+m}\P_{\mathcal{T}}(\{Y=k\} \cap \{X=x\}) + \\ 
        &\frac{m}{n+m}\P_{\mathcal{T}'}(\{Y=k\} \cap \{X=x\})  \\
        &= \frac{n}{n+m}\P_{\mathcal{T}}(X=x|Y=k)\P_{\mathcal{T}}(Y=k) + \\ 
        &\frac{m}{n+m}\P_{\mathcal{T}'}(X=x|Y=k)\P_{\mathcal{T}'}(Y=k) \\
        &= \left(\frac{n}{n+m}\P(Y=k)+\frac{m}{n+m}w_k\right)\P(X=x|Y=k) \\
        &= \P_{a}(Y=k)\frac{\P(Y=k|x)\P(x)}{\P(Y=k)}.
    \end{split}
\end{equation*}
By 
replacing this expression in 
\begin{equation*}
        \P_{a}(Y=k|x) = \frac{\P_{a}(\{Y=k\} \cap \{X=x\})}{\sum_{k}\P_{a}(\{Y=k\} \cap \{X=x\})},
\end{equation*}
conclude that
\begin{equation*}
    \begin{split}
        \P_{a}(Y=k|x) &= \frac{\P_{a}(\{Y=k\} \cap \{X=x\})}{\sum_{k}\P_{a}(\{Y=k\} \cap \{X=x\})} \\
        &= \frac{\frac{\P_{a}(Y=k)}{\P(Y=k)}\P(Y=k|x)\P(x)}{\sum_k \frac{\P_{a}(Y=k)}{\P(Y=k)}\P(Y=k|x)\P(x)} \\
        &= \frac{\frac{\P_{a}(Y=k)}{\P(Y=k)}\P(Y=k|x)}{\sum_k \frac{\P_{a}(Y=k)}{\P(Y=k)}\P(Y=k|x)}
    \end{split}
\end{equation*}
\end{proof}

\begin{proof}{Theorem 4}

By putting together Equations (1) and (2), we conclude that
$$g^*(\x):= \arg \min_{j \in \mathcal{Y}} \sum_{k \in \mathcal{Y}}  L'(k,j) \frac{\P(Y=k|\x)}{h(\x)},$$
where $h(\x)=\sum_{k=1}^K \frac{\P_{\a}(Y=k)}{\P(Y=k)}\P(Y=k|\x)$. The conclusion follows from the fact that $h$ is constant in $j$ and $k$.
\end{proof}

\begin{proof}[Corollary 1]

In the binary case, with an optimal classifier $g^*$ according to the induced probability and 0-1 loss. In the case where $\P_a(Y=1) = 0.5$, the loss function that gives the same classifier on the original probability is 
$$L'(1, 0)=\frac{\P_{\a}(Y=1)}{\P(Y=1)}=\frac{1}{2\P(Y=1)}$$ 
and 
$$L'(0, 1)=\frac{\P_{\a}(Y=0)}{\P(Y=0)}=\frac{1}{2\P(Y=0)}.$$

Based on Theorem 2, it follows that the decision rule corresponds is:

\begin{equation*}
    \begin{split}
        \frac{L'(0, 1)}{L'(0, 1) + L'(1, 0)} &= \frac{\frac{1}{2\P(Y=0)}}{\frac{1}{2\P(Y=0)}+\frac{1}{2\P(Y=1)}} \\
        &= \frac{\frac{1}{2\P(Y=0)}}{\frac{\P(Y=1)+\P(Y=0)}{2\P(Y=0)\P(Y=1)}} \\
        &= \frac{2\P(Y=0)\P(Y=1)}{2\P(Y=0)} \\
        &= \P(Y=1)
    \end{split}
\end{equation*}

To demonstrate that this classifier is the one that maximizes the balanced accuracy, we use the relation \citep{izbicki_mendonca2020} of the classifier $$g^*(x) = \I\left(\P(Y=1|\X)>\frac{l_1}{l_1+l_0}\right)$$ minimize the function risk $$R(g^*) = \E[l_1\I(Y=0, g^*(\X)=1) + l_0 \I(Y=1, g^*(\X)=0)].$$ 

Replacing $l_1=L'(0,1) = \frac{1}{\P(Y=0)}$ e $l_0=L'(1,0) = \frac{1}{\P(Y=1)}$ we have the same classifier than Criteria 5. We need to show that when minimizing this risk we obtained the best balanced accuracy. 
\begin{equation*}
    \begin{split}
        R(g^*) &= \E[L'(0,1)\I(Y=0, g^*(\X)=1) + L'(1,0) \I(Y=1, g^*(\X)=0)]\\
            &=L'(0,1)\P(Y=0, g^*(\X)=1) + L'(1,0)\P(Y=1, g^*(\X)=0) \\
            &=\frac{\P(Y=0, g^*(\X)=1)}{\P(Y=0)} + \frac{\P(Y=1, g^*(\X)=0)}{\P(Y=1)}\\
            &=\P(g^*(\X)=1|Y=0) + \P(g^*(\X)=0|Y=1)
    \end{split}
\end{equation*}

When we minimize this risk function we maximize the $$\P(g^*(\X)=1|Y=1) + \P(g^*(\X)=0|Y=0),$$ by definition this is the balanced accuracy.

\end{proof}

\section{Dataset Description}
\label{sec:sup_dataset}

Table \ref{tab:dataset} presents a description of the datasets utilized in our experiment. The table includes information regarding the input text, label variable, primary objective of the dataset, the data process, as well as the resultant proportion of labels.

\begin{sidewaystable}[htbp]
\caption{Description of the datasets used to compare the augmentation methods.}
\label{tab:dataset}
\begin{adjustbox}{scale=0.65}
\begin{tabular}{|l|l|l|l|l|l|}
    \hline
     Dataset Name & Goal & Text & Label & Label used on experiment & Ratio \\
     \hline
     \specialcell{Sentiment Analysis \\ in Twitter \\ \citep{barbieri2020tweeteval}} & \specialcell{Classifies a tweet sentiment \\ in negative, positive or neutral} & \textbf{text}: tweets to labeled & \specialcell{\textbf{label}: 0 if the sentiment \\is negative, 1 if is neutral, \\and 2, if it is positive.} & \specialcell{We used only the negative \\and positive tweets.} & $28\%$ of the tweets are negative\\
     \hline
     
     \specialcell{Women's E-Commerce \\ Clothing Reviews \\ \citep{agarap2018statistical} } & \specialcell{Based on a review classify \\ if the user recommends the \\product} &
     \specialcell{\textbf{Review text}: the text \\with the  product review} & \specialcell{\textbf{Recommend ID}: a binary \\ variable where 1 means that \\the person  recommends that \\product and 0 otherwise} & We used the data as it is. & \specialcell{18\% of the clients do not\\ recommend a product.}\\
     \hline
     
     \specialcell{Android Apps and \\ User Feedback \\ \citep{grano2017android}} & \specialcell{Classify an app review \\ in positive or negative} & \specialcell{\textbf{review}: the user review\\ of a software application \\package} & \specialcell{\textbf{star}: user rate, from 1 to 5, \\where 5 means a good review} & \specialcell{The reviews with stars \\greater than 3 we consider \\ a positive, and less than \\ this a negative.} & 18\% of the reviews are negative \\
     \hline
     
     \specialcell{Yelp reviews \\ \citep{zhang2015character}} & \specialcell{Classify if a yelp review \\ is a good or a bad review} & \textbf{text}: review texts & \specialcell{\textbf{label}: user rate, from 1 to 5, \\ where 5 means a good review} & \specialcell{The reviews with stars \\greater than 3 we consider \\a positive, and less than \\this a negative.} & \specialcell{As the dataset is balanced, we \\took a sample with only 10\% \\of negative reviews.} \\
     
     \hline
     \specialcell{Multilingual Amazon \\ review corpus \\ \citep{keung2020multilingual}} & \specialcell{Classify an Amazon product \\ review is positive or negative} &\specialcell{\textbf{product review}: client \\ review over an amazon \\product} & \specialcell{\textbf{stars}:  user rate, from 1 to 5, \\ where 5 means a good review} & \specialcell{The reviews with stars \\greater than 3 we consider \\a positive, and less than \\this a negative.} & \specialcell{As the dataset is balanced, we\\ took a  sample with only 10\% of \\negative reviews.} \\
     \hline
     
     \specialcell{Hate speech offensive \\ \citep{davidson2017automated}} & \specialcell{Label a tweet as hate speech,\\ offensive or neither} & \textbf{tweet}: tweets to be labeled & \specialcell{\textbf{class}: 0 if the tweet is hate \\speech,  1 if it is offensive, \\and 2 neither} &  \specialcell{We only use classes hate \\speech and offensive tweets}  & \specialcell{7\% of the tweets are labeled as \\hate speech.}\\
     \hline
     \specialcell{Jigsaw toxicity dataset \\ \citep{manerba2022bias}} & \specialcell{Classify if a comment made in \\ the Civil Comments platform \\ is toxic or not, and classify in \\sub-types of toxicity} &\specialcell{\textbf{comment\_text}: comments \\ made in the Civil Comments \\platform} & \specialcell{\textbf{target}: human rate that \\thinks that the comment \\is toxic} & \specialcell{We used comment \\with rate greater than \\0.5 as toxic.} & 6\% of the comments are toxic. \\
     \hline
     \specialcell{Jigsaw toxicity dataset \\ \citep{manerba2022bias}} & \specialcell{Classify if a comment made in \\ the Civil Comments platform \\ is toxic or not, and classify in \\sub-types of toxicity} &  \specialcell{\textbf{comment\_text}: comments \\ made in the Civil Comments \\platform} & \specialcell{\textbf{insult}: human rate that \\thinks that the comment is \\insulting (a sub-types of \\toxicity)}& \specialcell{ We used comment \\with rate greater than \\0.5 as insult} & 4\% of the comments are insulting.\\   
     \hline
\end{tabular}
\end{adjustbox}

\end{sidewaystable}

\section{ImpOrtance Word Augmentation (IOWA)}
\label{sec:iowa}

The Importance Word Augmentation is a method to generate new sentences based on the frequency of each word given in the class. To generate the new sentence, we need $\lambda_k$, the mean of the length of each sentence of label $k$, and $W$, a matrix of dimension $D \times V$, where D is the number of sentences, on the train set, and V is the number of the words in the training vocabulary. Then, $W$ is filled with the frequency of each word in each sentence, so $w_{i,j}$ is the frequency of word $j$ in sentence $i$. 

To generate the new sample of a class $k$,  the following rules are used:
\begin{enumerate}
    \item Generate N, the length of the sentence, from $Poisson(\lambda_k)$
    \item Generate the vector X, words of the sentence, from a $Multinomial(N, g(W,k, y))$,
\end{enumerate}
where $g(W, k, y)$ is a weight function to generate the probability based on $W$. Next, we define four alternatives to construct this function. 

\subsection{IOWA using Frequency}
The first method is using only the frequency of the words, $$g(W, k, y) = \left[\frac{\sum_{i=1}^D w_{i1}I_{y_i=k}}{\delta_k}, \ldots, \frac{\sum_{i=1}^D w_{iV}I_{y_i=k}}{\delta_k}\right],$$ where $\delta_k = \sum_{j=1}^V\sum_{i=1}^Dw_{ij}I_{y_i = k}$.

This method will assign higher probabilities for words that are frequent in sentences of the label $k$.

\subsection{IOWA using the Difference of the frequencies}

In this method we generate the frequency of a word for each label (similar to the first method) and compute the difference between the frequency on the desired label and the others, this is made to give extra weight to more frequent words in the desired label. We will break the construction of $g(W, k, y)$ into small steps:
\begin{enumerate}
    \item First generate for each word the relative frequency in respect to the desired label ($k$), $$s_{j,k} = \frac{\sum_{i=1}^D w_{ij}I_{y_i=k}}{\sum_{i=1}^DI_{y_i=k}}$$
    \item Then generate from the others $$s_{j,k}^c = \frac{\sum_{i=1}^D w_{ij}I_{y_i\neq k}}{\sum_{i=1}^DI_{y_i\neq k}}$$
    \item Calculate the difference of the frequency for each word $$d_{j, k} = s_{j, k} - s_{j, k}^c$$
    \item $\delta_k = \sum_{j=1}^{V} d_{j,k} I_{d_{j,k} > 0}$
    \item And finally $$g(W, k, y) = \left[\frac{d_{j,k} I_{d_{j,k} > 0}}{\delta_k}, \ldots , \frac{d_{V,k} I_{d_{V,k} > 0}}{\delta_k} \right]$$
\end{enumerate}

This method is similar to the IOWA using Frequency, but this time we will assign higher probabilities for the more frequent words in sentences of the label $k$.

\subsection{IOWA using the Inverse Document Frequency (IDF)}
Another possibility is to generate the vector with the inverse document frequency (IDF). This method will give more rare words greater frequency, the generation of this method is similar to the frequency, but we will need to obtain the IDF of the words. 

The IDF of a word (with respect to the label $k$) is given by $$IDF_{j, k} = \log\left(\frac{\sum_{i=1}^DI_{y_i=k}}{\sum_{i=1}^Dw_{ij}I_{y_i=k}}\right).$$ And $g(W, k, y)$ is given by: $$g(W, k, y) = \left[\frac{IDF_{1, k}}{\delta_k}, \ldots, \frac{IDF_{V, k}}{\delta_k}\right],$$ where $\delta_k = \sum_{j=1}^V IDF_{j, k}$

\subsection{IOWA using the Difference of the IDF}
The last method takes the difference between the IDF of the words constructed on the reference label and the IDF in other labels. To make this we have to calculate $$IDF_{j, k}^c = \log \left(\frac{\sum_{i=1}^D I_{y_i\neq k}}{\sum_{i=1}^Dw_{ij}I_{y_i \neq k}}\right),$$ and then calculate the difference $d_j = IDF_{j, k} - IDF_{j, k}^c$, so $$g(W, k, y) = \left[\frac{d_1 I_{d_1 > 0}}{\delta_k}, \ldots, \frac{d_V I_{d_V > 0}}{\delta_k} \right],$$ where $\delta_k = \sum_{j=1}^V d_j I_{d_j > 0}$.

\section{Hypothesis test}
\label{sec:hypothesis}
In this section, we present the hypothesis text adopted in the main manuscript to check the difference in the performance of the original and augmented databases.

Fix an augmentation method and a dataset, and
let $X_{i,j}$ be the percentage gain on the 
$j$-th augmented dataset for the $i$-th sample of the original data, $i=1,\ldots,5$ and $i=1,\ldots,J$.
In order to take the dependency between measurements
obtained on the same sample of the original data, 
we assume that
$X_{i,j} \sim N(M_i,\sigma^2_R)$ are 
independent random variables given $M_1,\ldots,M_5$, that 
$M_1,\ldots,M_5 \overset{\text{i.i.d.}}{\sim}  N(\mu,\sigma^2)$, and that $\mu,\sigma^2,\sigma^2_R$ are fixed parameters. 

Our goal is to test the null hypothesis $H_0:\mu=0$ vs $H_1:\mu\neq 0$.
First, we compute the test statistic $T=\frac{1}{5J}\sum_{i,j}X_{i,j}$, the average percentage gain. Then, we compute p-values based on $T$ via a parametric bootstrap. This is done by first
estimating $\sigma^2$ and $\sigma^2_R$ using their maximum likelihood estimates, $\widehat \sigma^2,\widehat  \sigma^2_R$. Then, we sample the test statistic from the null by sample data with the same structure as $X_{i,j}$'s at the point $(\mu,\sigma^2,\sigma^2_R)=(0,\widehat \sigma^2,\widehat  \sigma^2_R)$ and computing the test statistic for each sampled dataset, $T^{(1)},\ldots,T^{(B)}$.
The p-value is simply
$$\frac{1}{B}\sum_{b=1}^B I\left(|T^{(b)}|\geq |T|\right).$$
We take $B=1000$.

\section{Logistic Analysis}
\begin{figure}[!h]
    \centering
    \includegraphics[width=.9\textwidth]{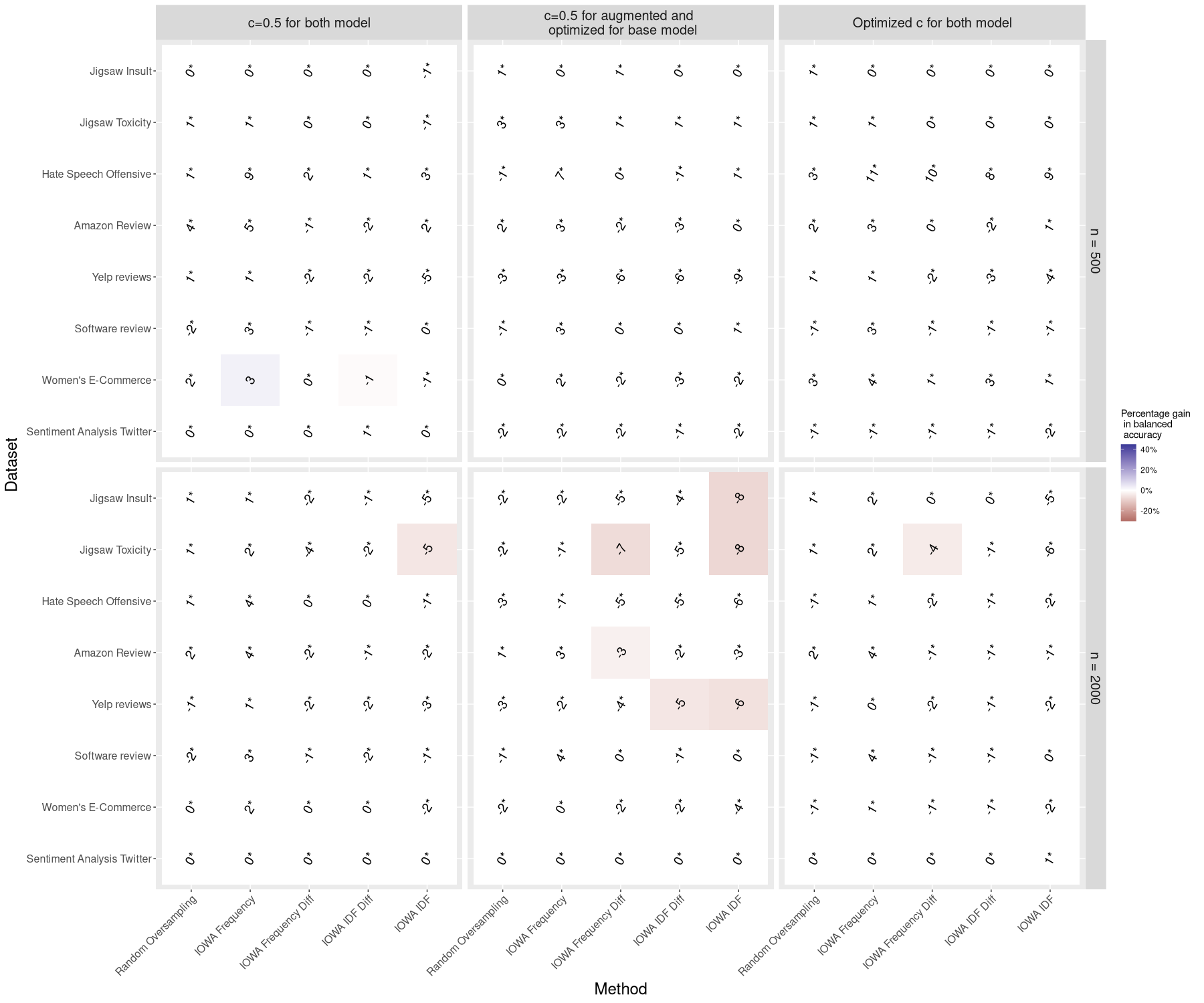}
    \caption{Heatmap of the mean percentage gain in balanced accuracy when comparing the augmented methods with the non-augmented model for classification rules.  Positive values indicate superior performance by the augmented method. Non-significant gains are marked with asterisks and displayed in white. Our findings indicate that for the three comparisons of threshold choice the gain is non-significant.}
    \label{fig:log_heatmap}
\end{figure}

In this section, we present the results of training the IOWA and Random Oversampling method with a logistic model and compare the model with the same comparative criteria presented in Section 3. Figure \ref{fig:log_heatmap} displays the percentage gain in balanced accuracy for the logistic regression. Almost all experiments did not result in any gain in accuracy. 

\label{sec:log}

\begin{figure}[!h]
    \centering
    \includegraphics[width=.9\textwidth]{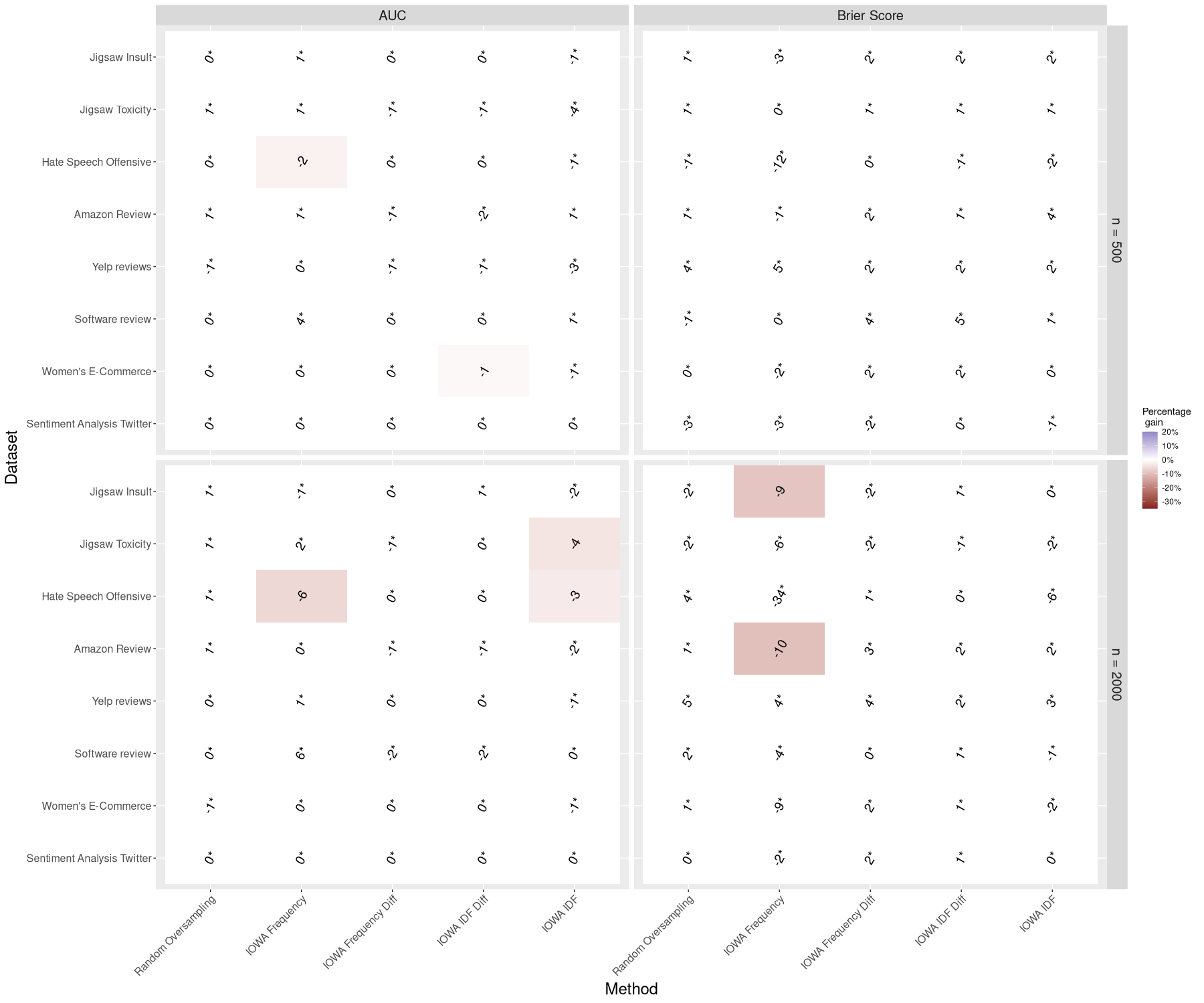}
    \caption{Heatmap of the average percentage improvement in the AUC (left column) and Brier Score (right column) when comparing the augmented methods with the non-augmented ones.  Positive values indicate superior precision in estimating $\P(Y=1|\x)$ using the augmented method. Non-significant gains are marked with asterisks and displayed in white. The results indicate that data augmentation never improves $\P(Y=1|\x)$ estimates.}
    \label{fig:log_metrics}
\end{figure}

Figure \ref{fig:log_metrics} shows the gain on the AUC and Brier Score. The results indicate that the estimation is better with the non-augmented method.

\section{ROC curve analysis}
\label{sec:roc_curve}

This section  presents the ROC curve analysis shown in Section 3 for the other datasets. The top row shows the settings with greater gains in the AUC, while  the bottom row the cases with the lowest gain. Overall the conclusions are similar to the ones discussed in the main manuscript.

\begin{figure}[htbp]
    \centering
    \includegraphics[width=.7\textwidth]{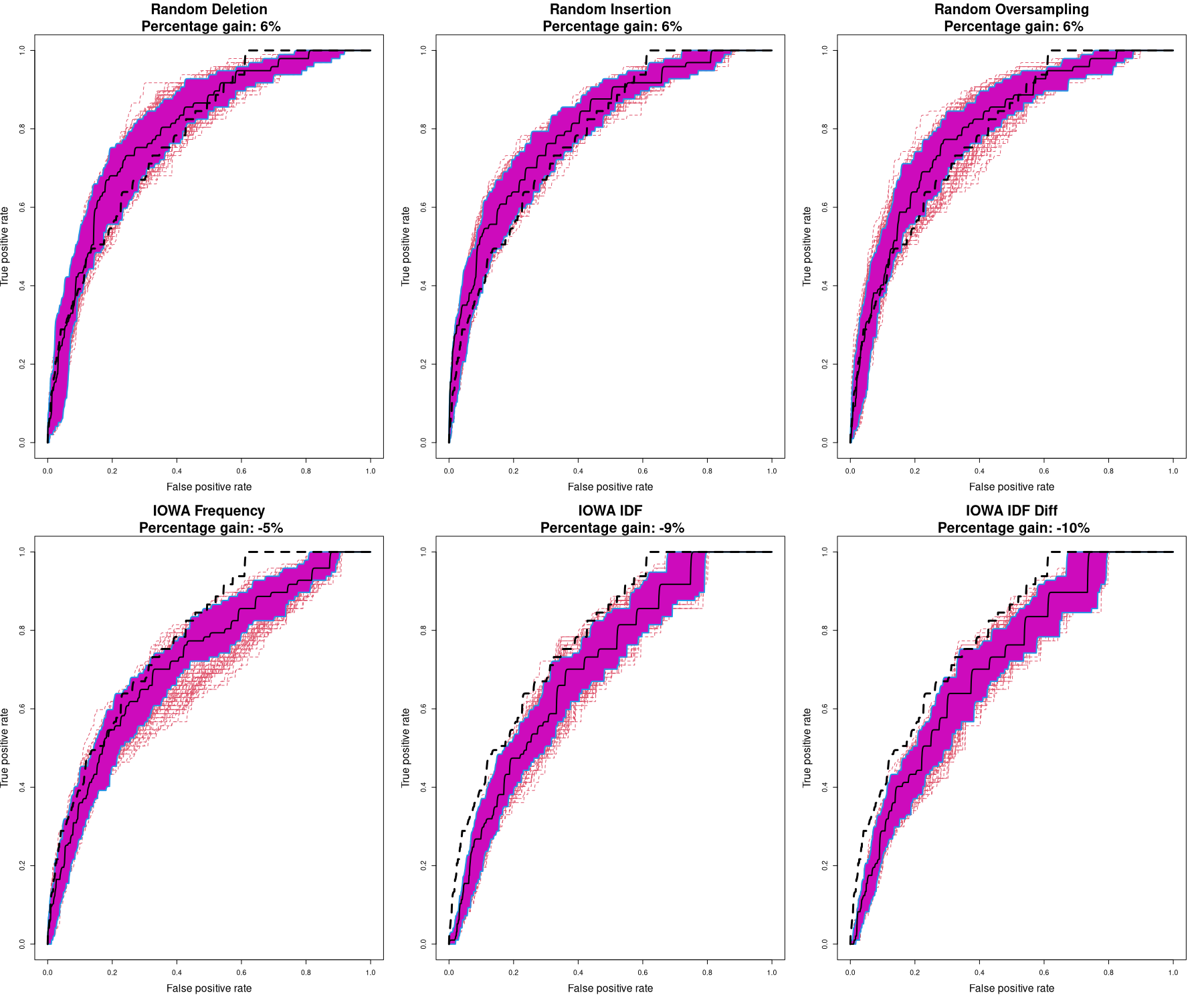}
    \caption{Functional BoxPlot of the Yelp dataset on the train size of 2000}
\end{figure}

\begin{figure}[htbp]
    \centering
    \includegraphics[width=.7\textwidth]{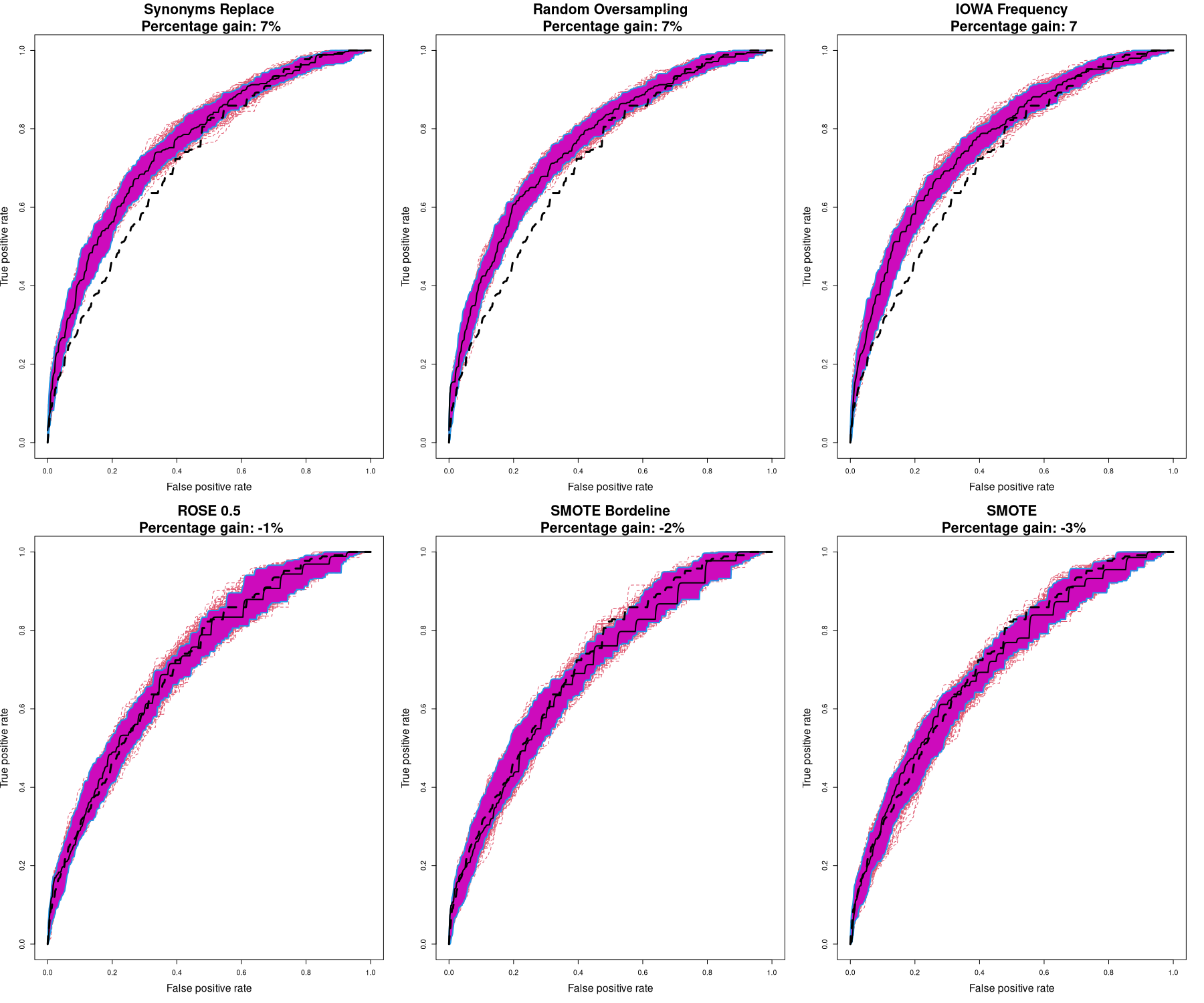}
    \caption{Functional BoxPlot of the Sentiment Twitter dataset on the train size of 500}
\end{figure}

\begin{figure}[htbp]
    \centering
    \includegraphics[width=.7\textwidth]{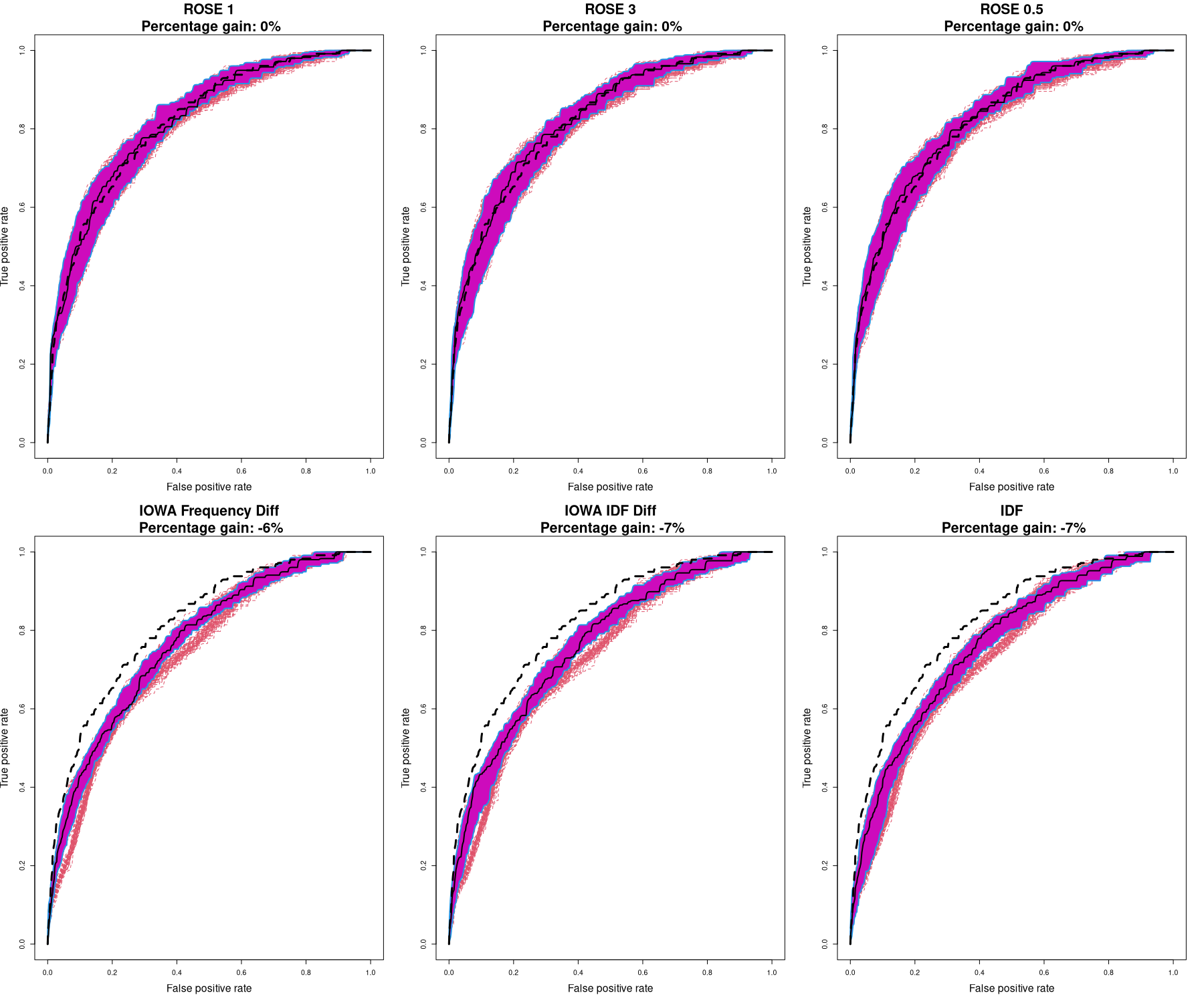}
    \caption{Functional BoxPlot of the Sentiment Twitter dataset on the train size of 2000}
\end{figure}

\begin{figure}[htbp]
    \centering
    \includegraphics[width=.7\textwidth]{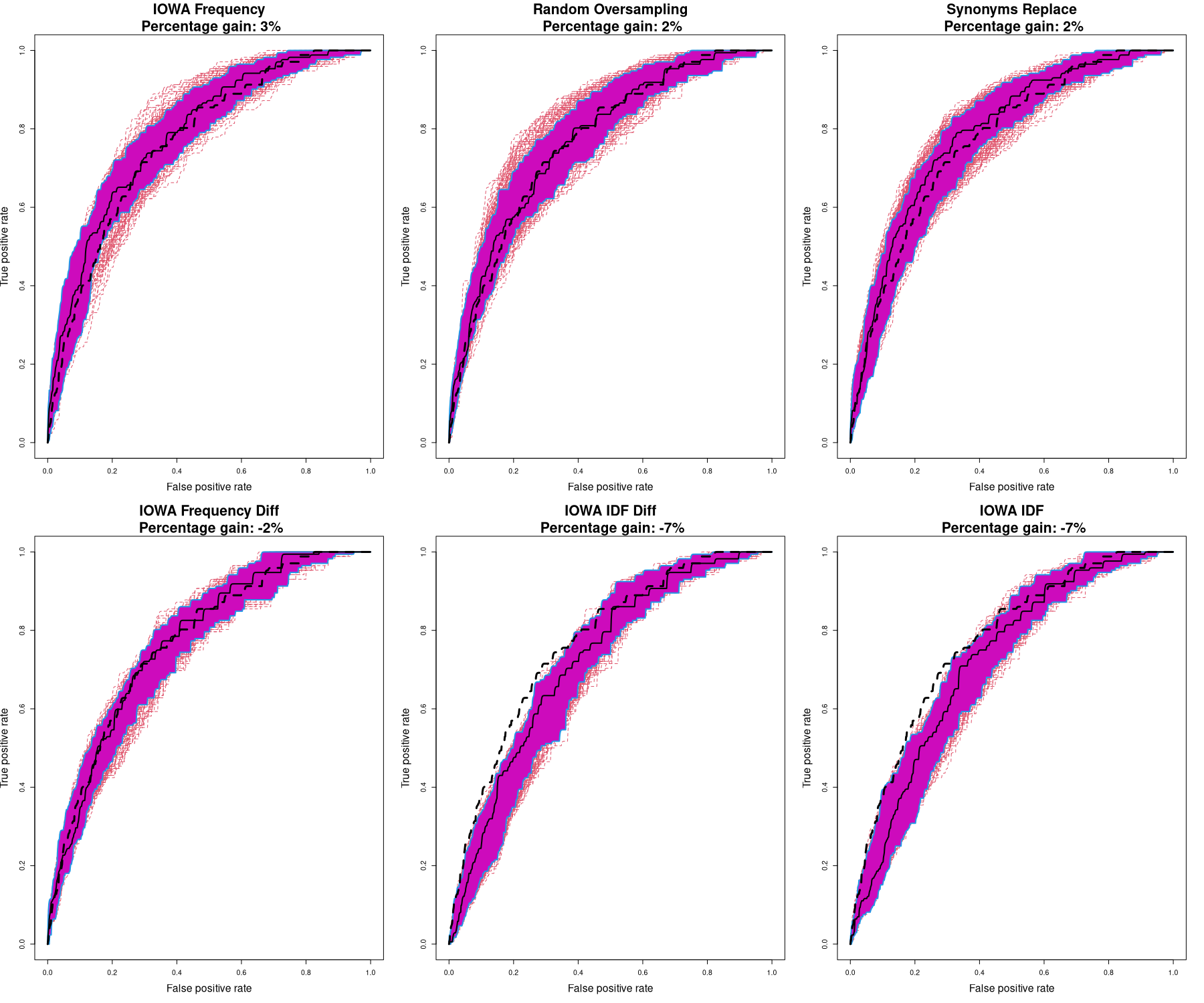}
    \caption{Functional BoxPlot of the Women's E-Commerce dataset on the train size of 500}
\end{figure}

\begin{figure}[htbp]
    \centering
    \includegraphics[width=.7\textwidth]{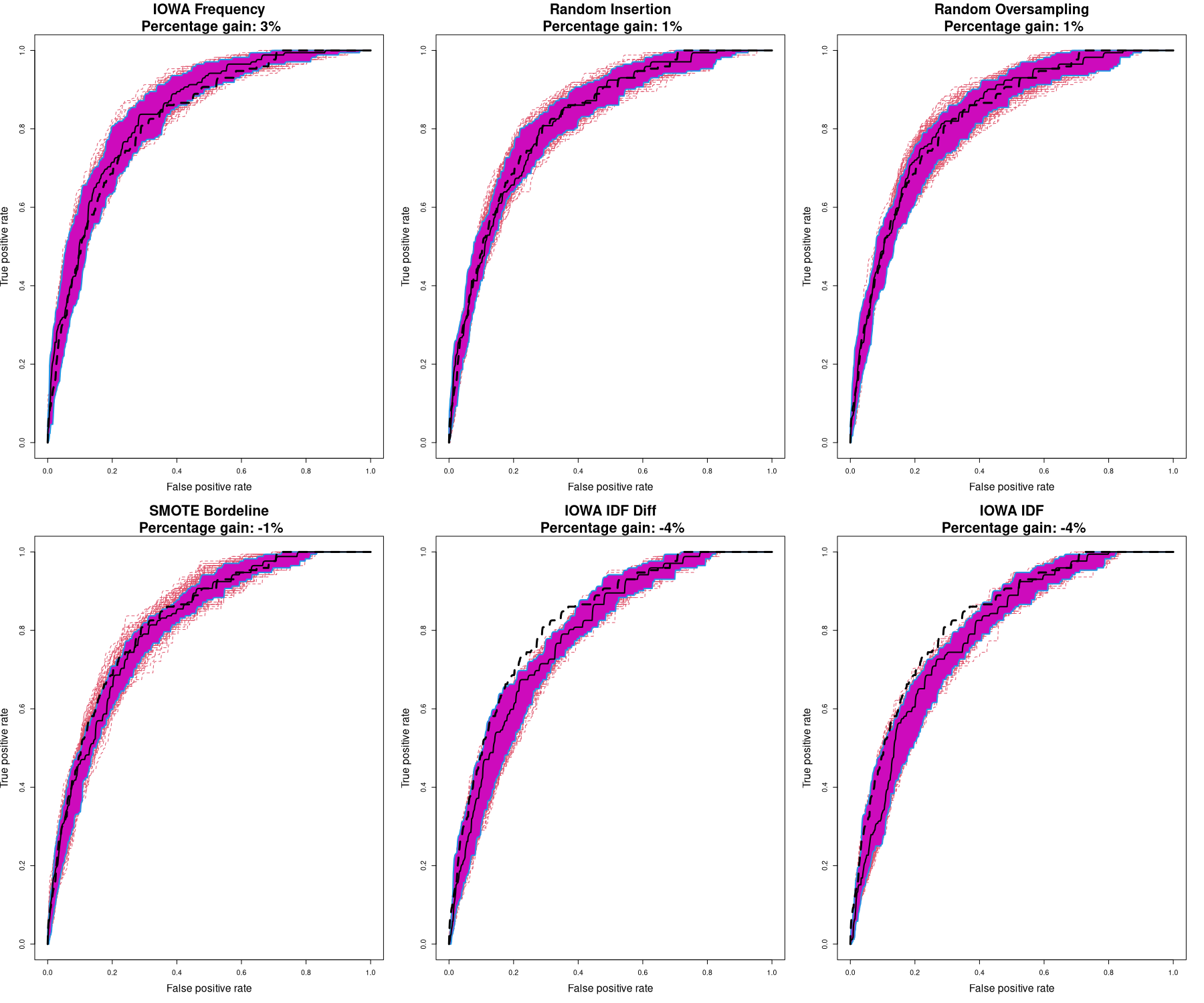}
    \caption{Functional BoxPlot of the Women's E-Commerce dataset on the train size of 2000}
\end{figure}

\begin{figure}[htbp]
    \centering
    \includegraphics[width=.7\textwidth]{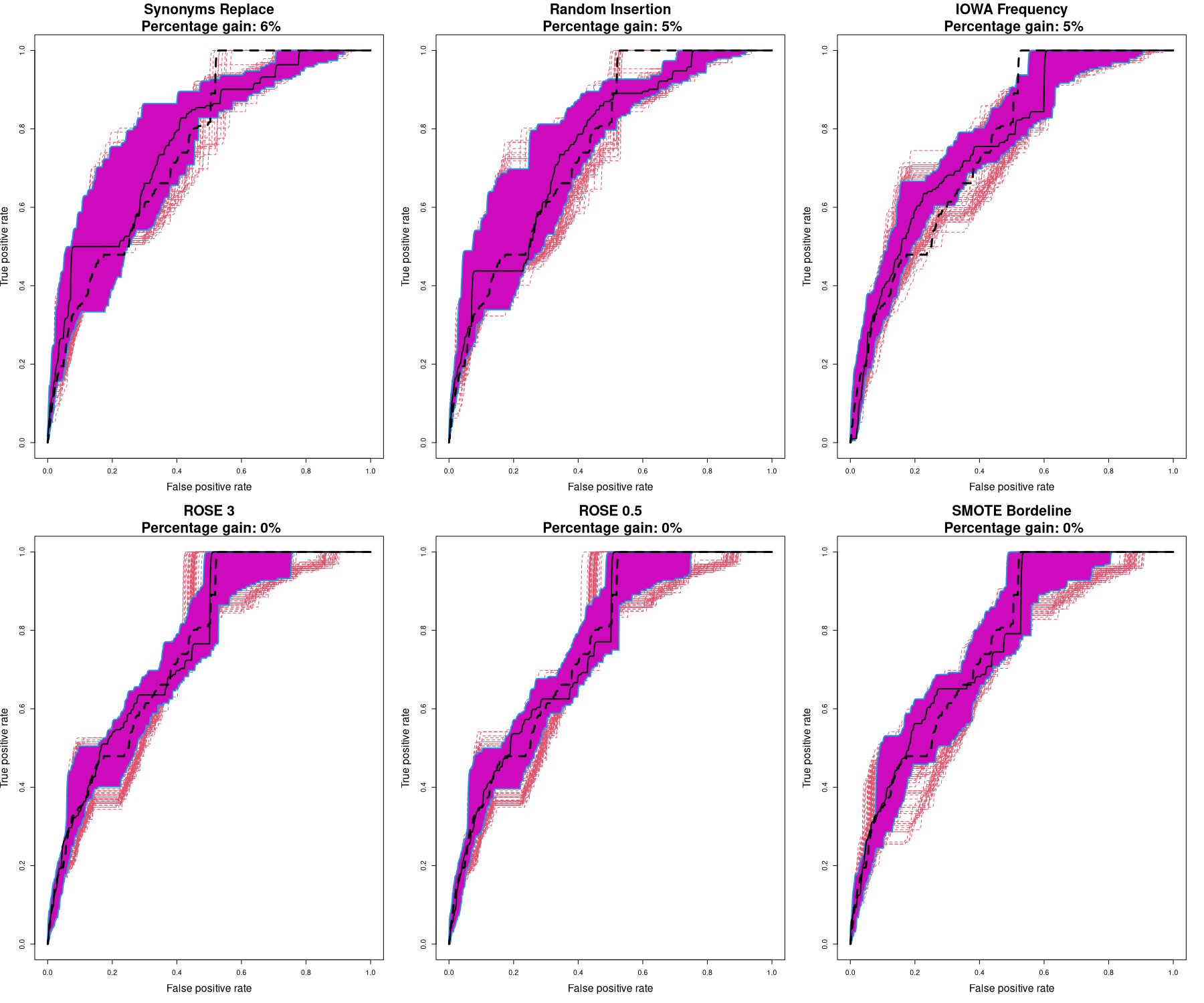}
    \caption{Functional BoxPlot of the Software review dataset on the train size of 500}
\end{figure}

\begin{figure}[htbp]
    \centering
    \includegraphics[width=.7\textwidth]{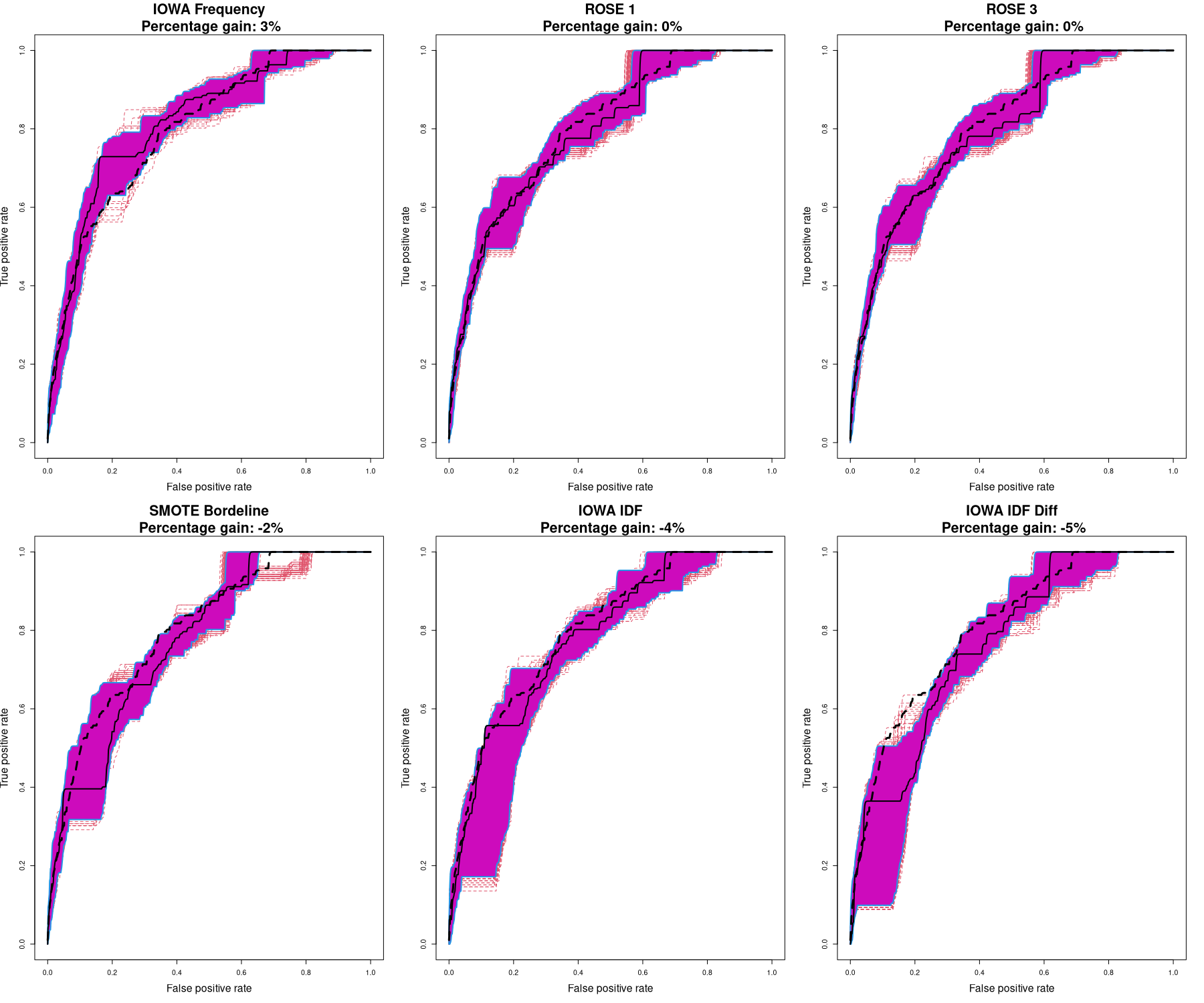}
    \caption{Functional BoxPlot of the Software review dataset on the train size of 2000}
\end{figure}

\begin{figure}[htbp]
    \centering
    \includegraphics[width=.7\textwidth]{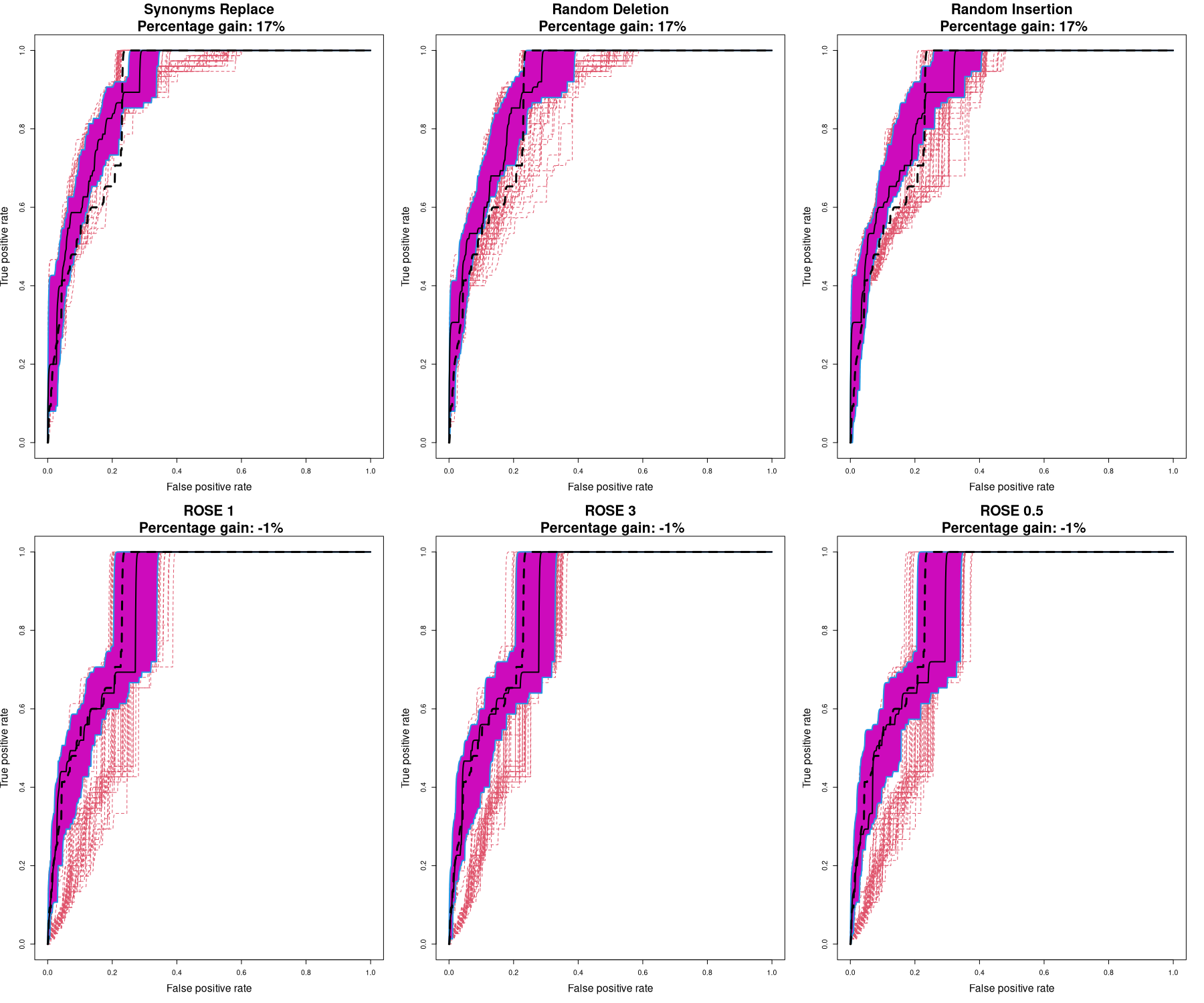}
    \caption{Functional BoxPlot of the Hate Speech Offensive dataset on the train size of 500}
\end{figure}

\begin{figure}[htbp]
    \centering
    \includegraphics[width=.7\textwidth]{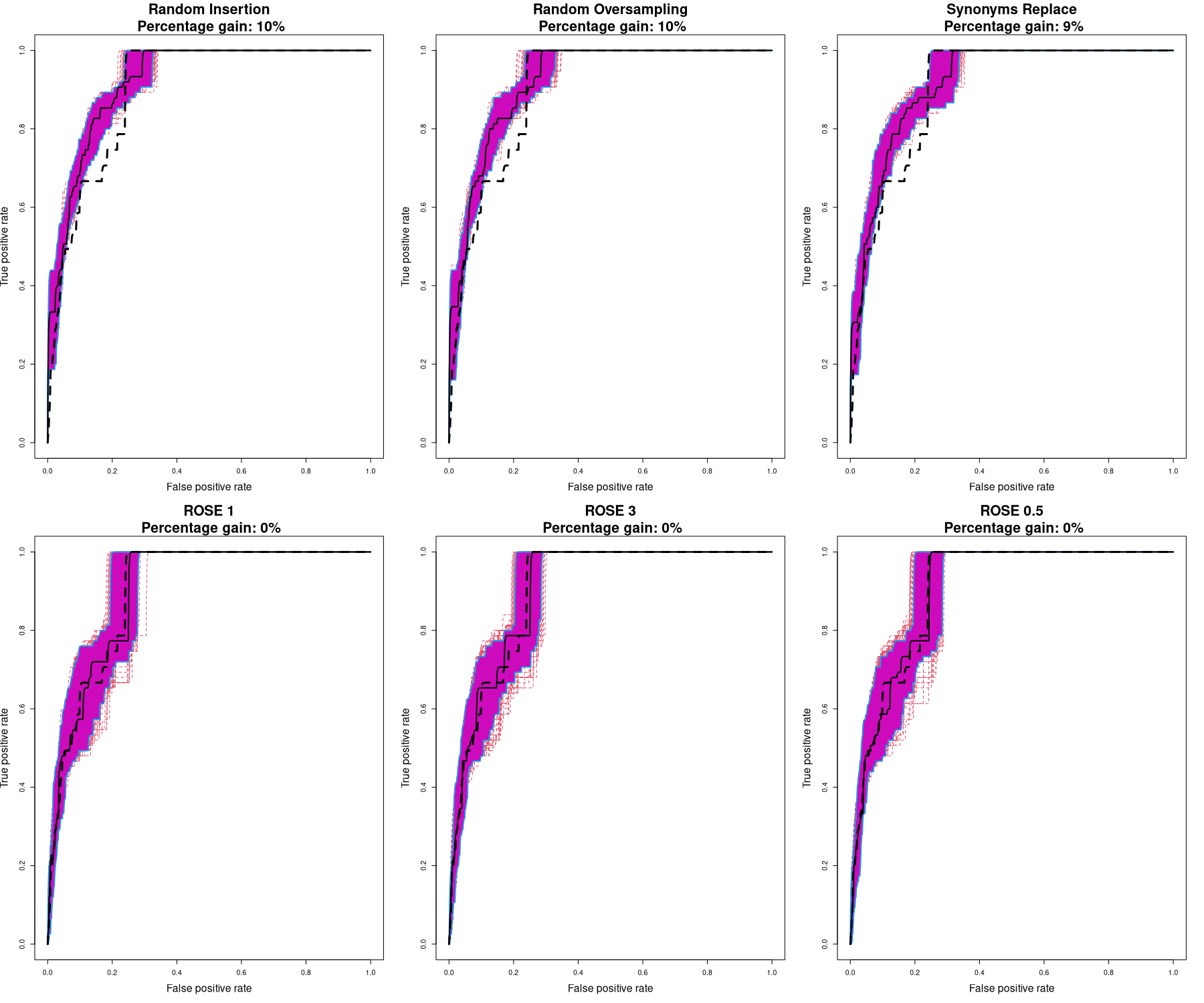}
    \caption{Functional BoxPlot of the Hate Speech Offensive dataset on the train size of 2000}
\end{figure}

\begin{figure}[htbp]
    \centering
    \includegraphics[width=.7\textwidth]{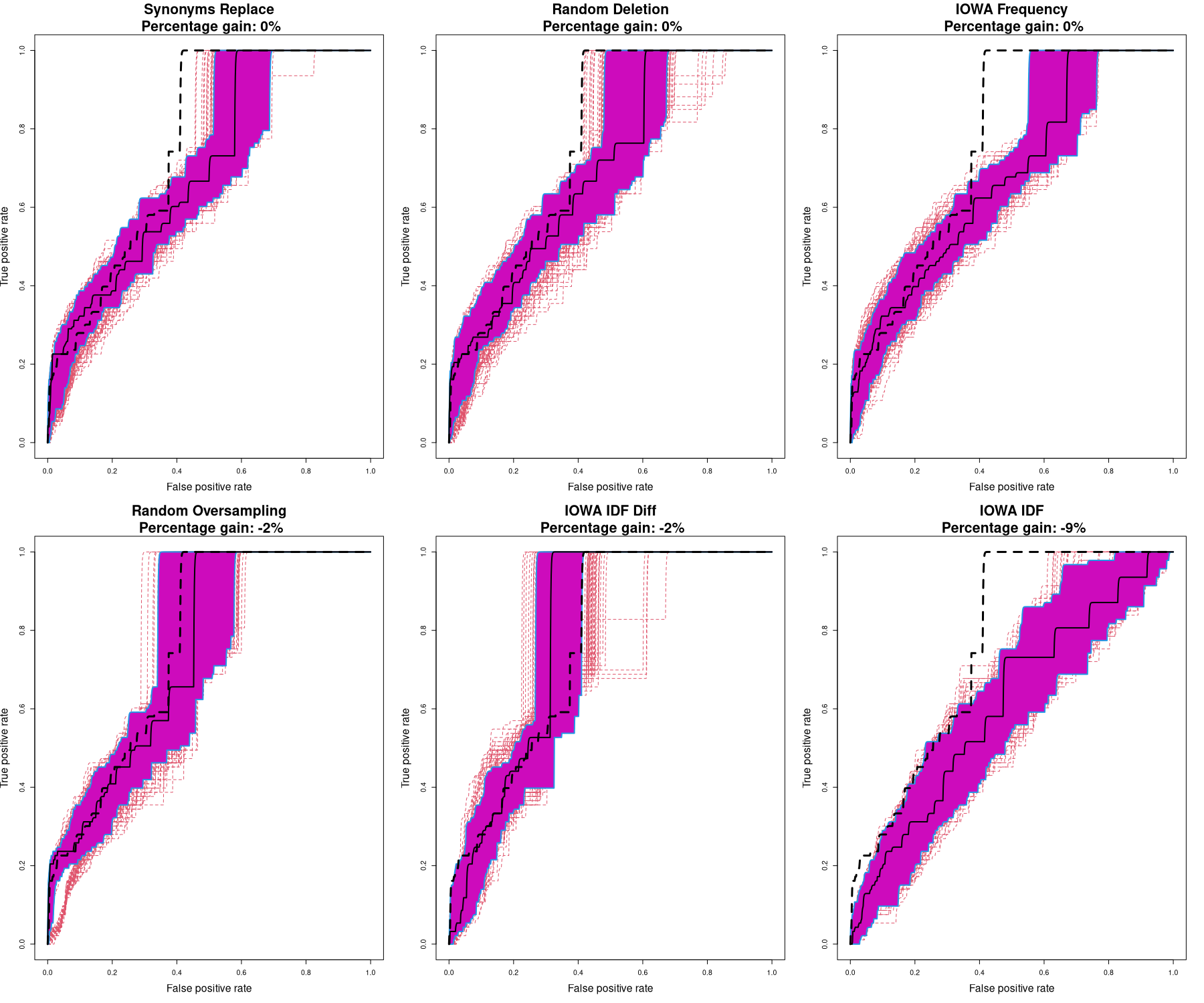}
    \caption{Functional BoxPlot of the Jigsaw Toxicity dataset on the train size of 500}
\end{figure}

\begin{figure}[htbp]
    \centering
    \includegraphics[width=.7\textwidth]{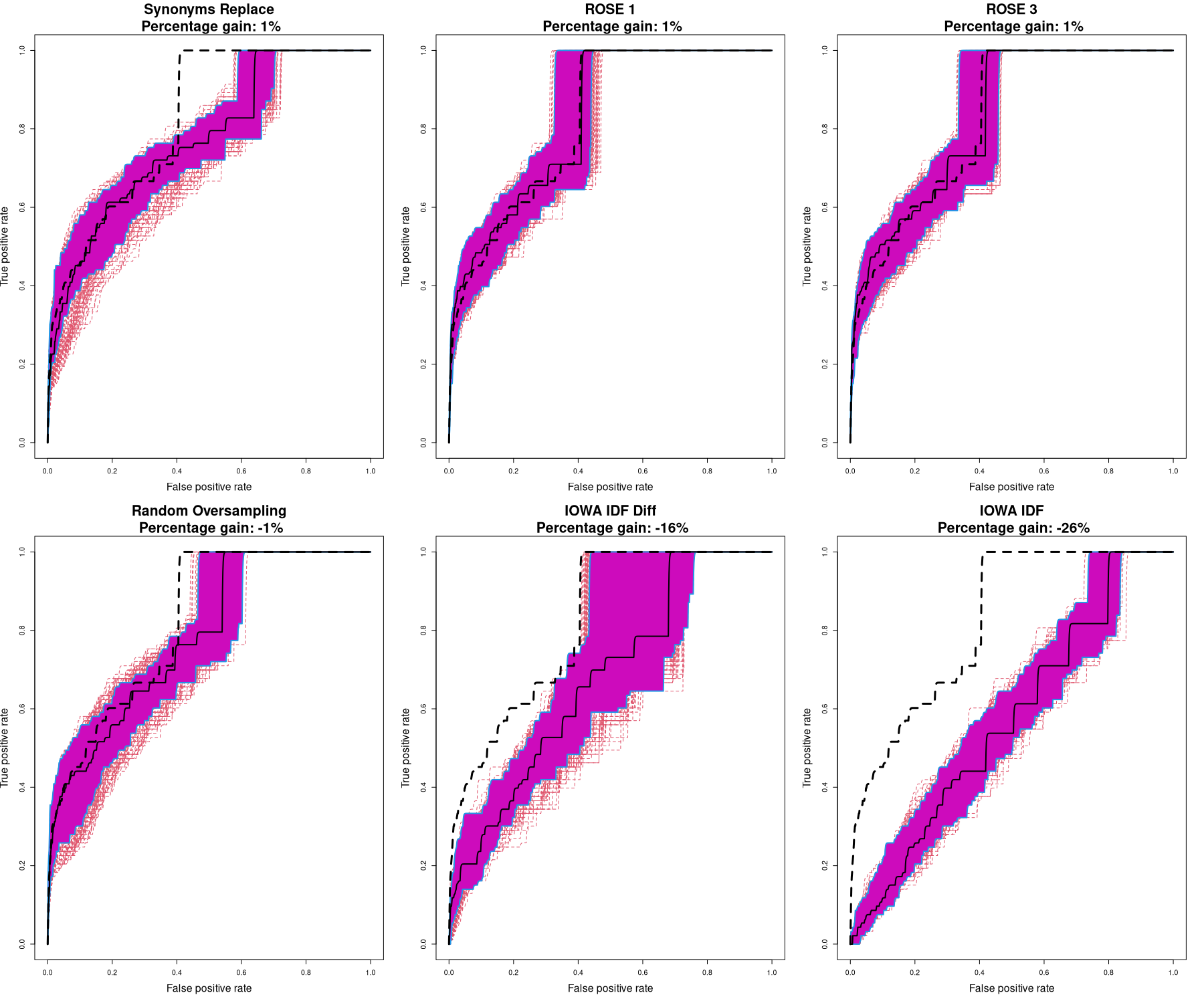}
    \caption{Functional BoxPlot of the Jigsaw Toxicity dataset on the train size of 2000}
\end{figure}

\begin{figure}[htbp]
    \centering
    \includegraphics[width=.7\textwidth]{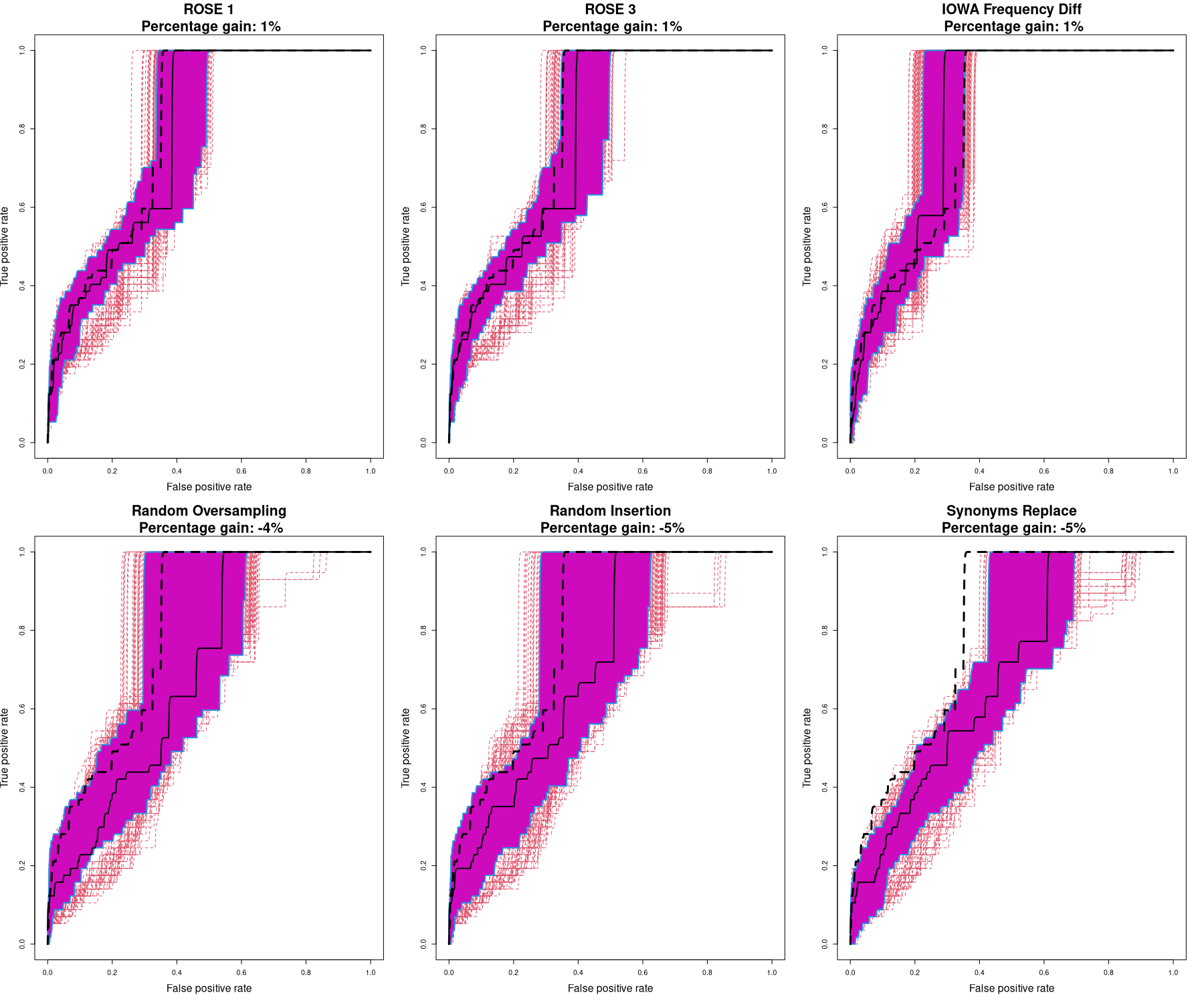}
    \caption{Functional BoxPlot of the Jigsaw Insult dataset on the train size of 500}
\end{figure}

\newpage
\begin{figure}[htbp]
    \centering
    \includegraphics[width=.7\textwidth]{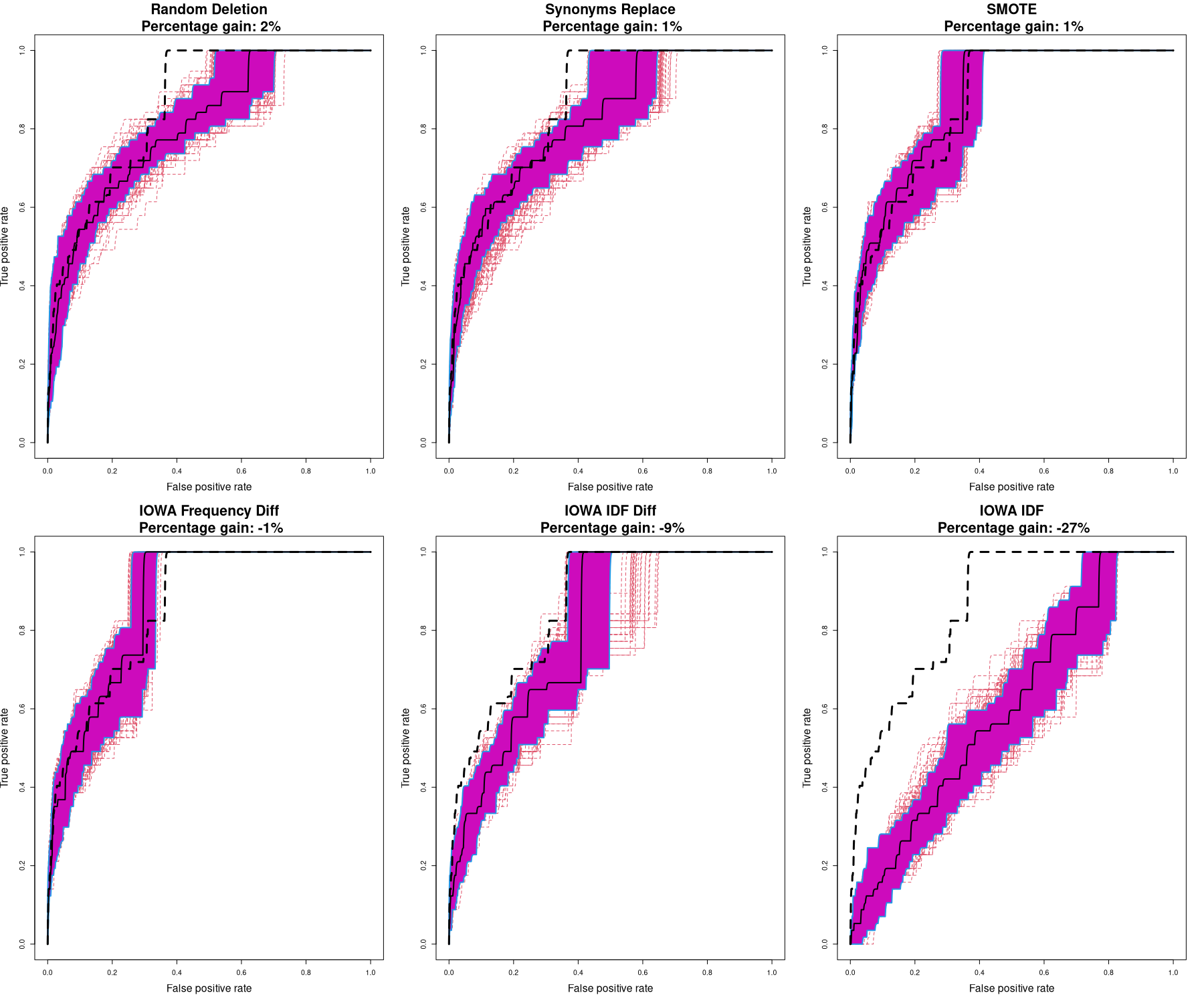}
    \caption{Functional BoxPlot of the Jigsaw Insult dataset on the train size of 2000}
\end{figure}

\section{Other metrics analysis}
\label{sec:other_met}
We evaluate the percentage gain over other metrics. In this section, we displayed the results for the F1-score, Accuracy, Sensitivity, and Specificity. The outcomes for F1-score and sensitivity exhibited a resemblance to those obtained for balanced accuracy. Although the default threshold produced a perceived enhancement, optimization of the threshold led to similar outcomes. In contrast, while utilizing the default threshold did not lead to improvements with augmented methods for specificity and accuracy, the optimized threshold showed an improvement in the augmented model, particularly in terms of specificity.

\begin{figure}[htbp]
    \centering
    \includegraphics[width=.9\textwidth]{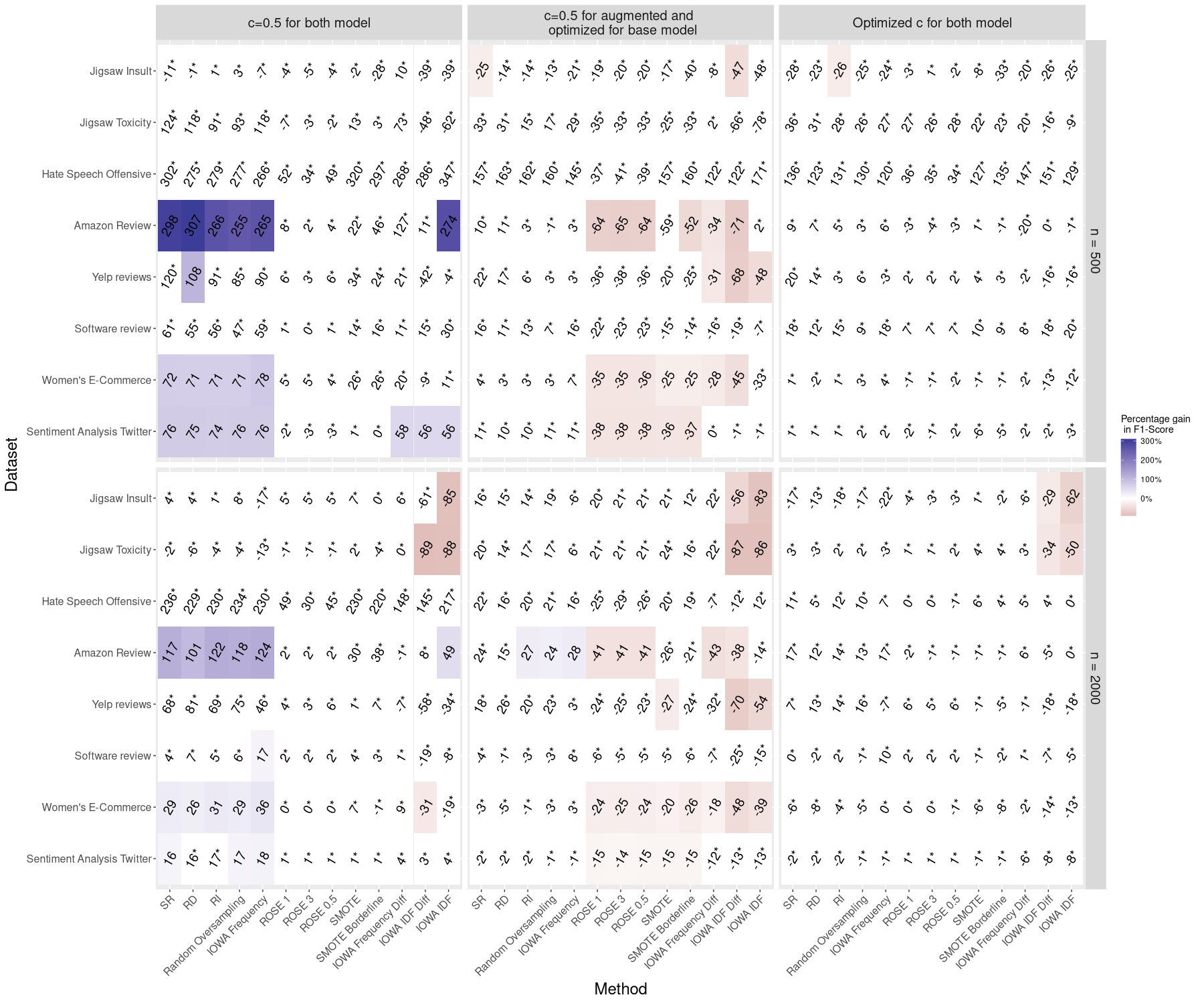}
    \caption{Heatmap of the mean percentage gain in F1-Score when comparing the augmented methods with the non-augmented model for classification rules.  Positive values indicate superior performance by the augmented method. Non-significant gains are marked with asterisks and displayed in white. Our findings indicate a similar result to the balanced accuracy, the data augmentation provides a noticeable benefit only when using the default threshold of $c=0.5$ (left column); optimizing the threshold on non-augmented data eliminates the need for augmentation.}
\end{figure}

\begin{figure}[htbp]
    \centering
    \includegraphics[width=.9\textwidth]{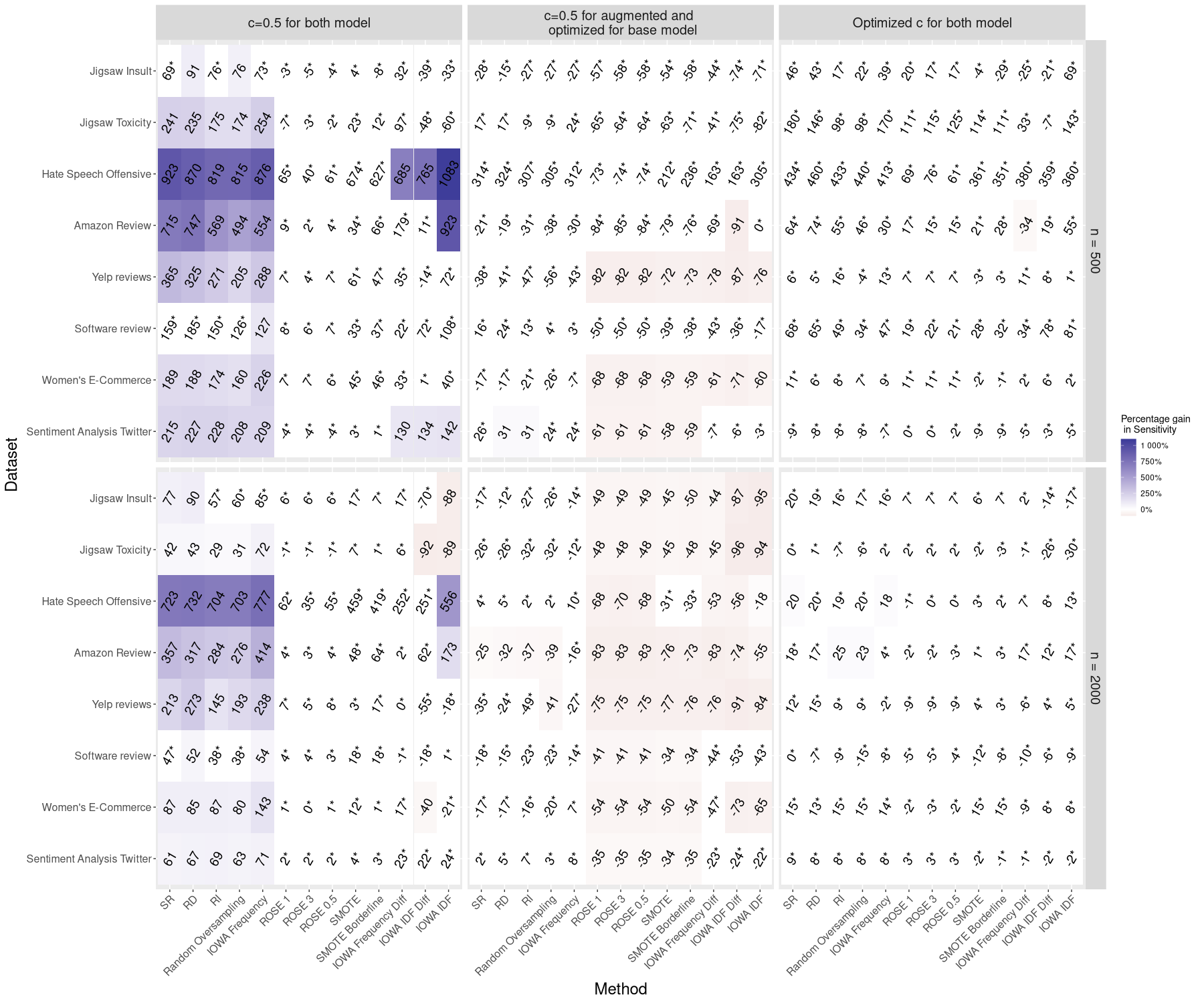}
    \caption{Heatmap of the mean percentage gain in Sensitivity (minority class) when comparing the augmented methods with the non-augmented model for classification rules.  Positive values indicate superior performance by the augmented method. Non-significant gains are marked with asterisks and displayed in white. Our findings for this indicate that optimizing the threshold on non-augmented eliminates the need for augmentation.}
\end{figure}

\begin{figure}[htbp]
    \centering
    \includegraphics[width=.9\textwidth]{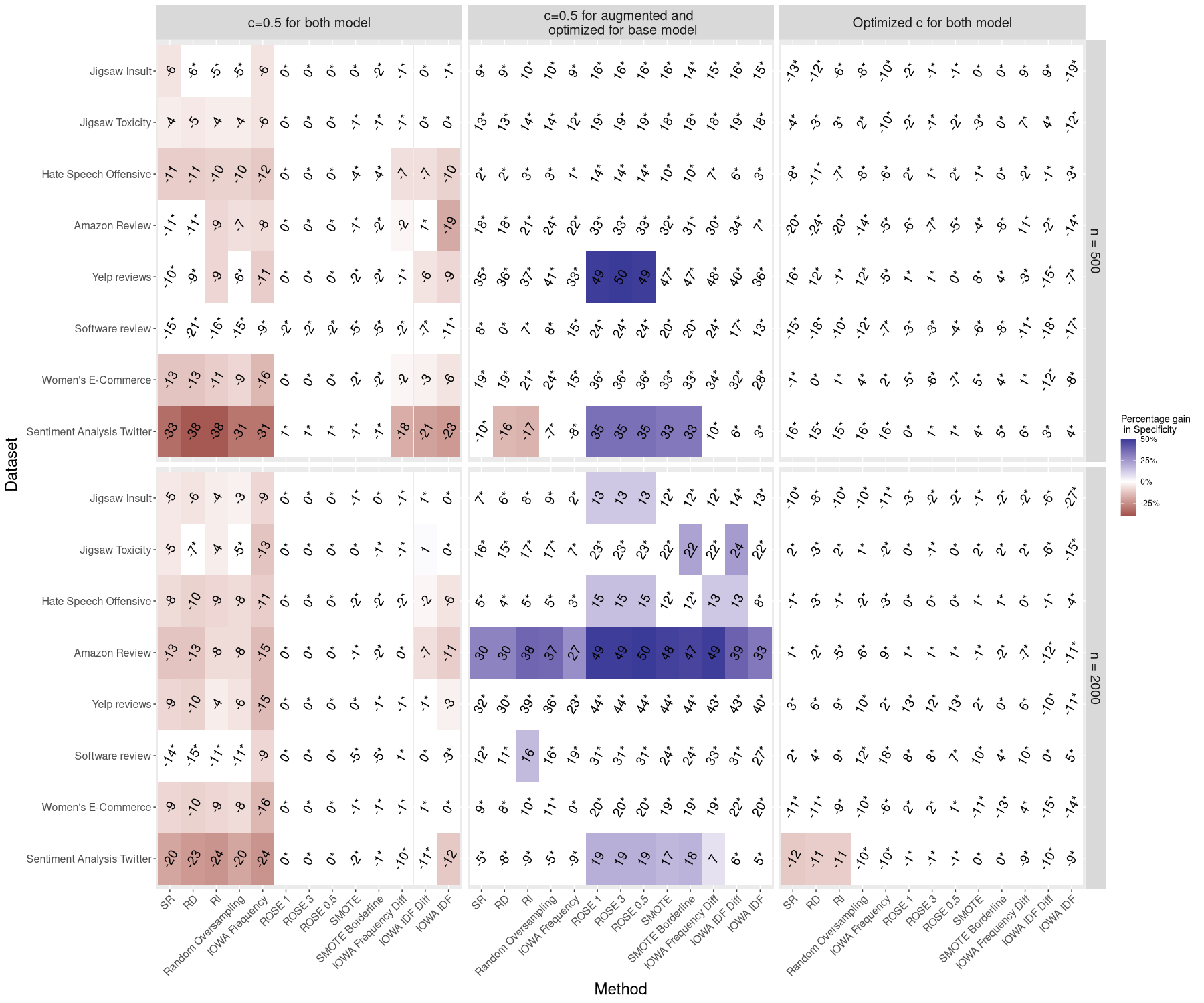}
    \caption{Heatmap of the mean percentage gain in Specificity (majority class) when comparing the augmented methods with the non-augmented model for classification rules.  Positive values indicate superior performance by the augmented method. Non-significant gains are marked with asterisks and displayed in white. Our findings indicate a different behavior for this metric, when using the optimized threshold the augmented method  shows an improvement.}
\end{figure}

\begin{figure}[htbp]
    \centering
    \includegraphics[width=.9\textwidth]{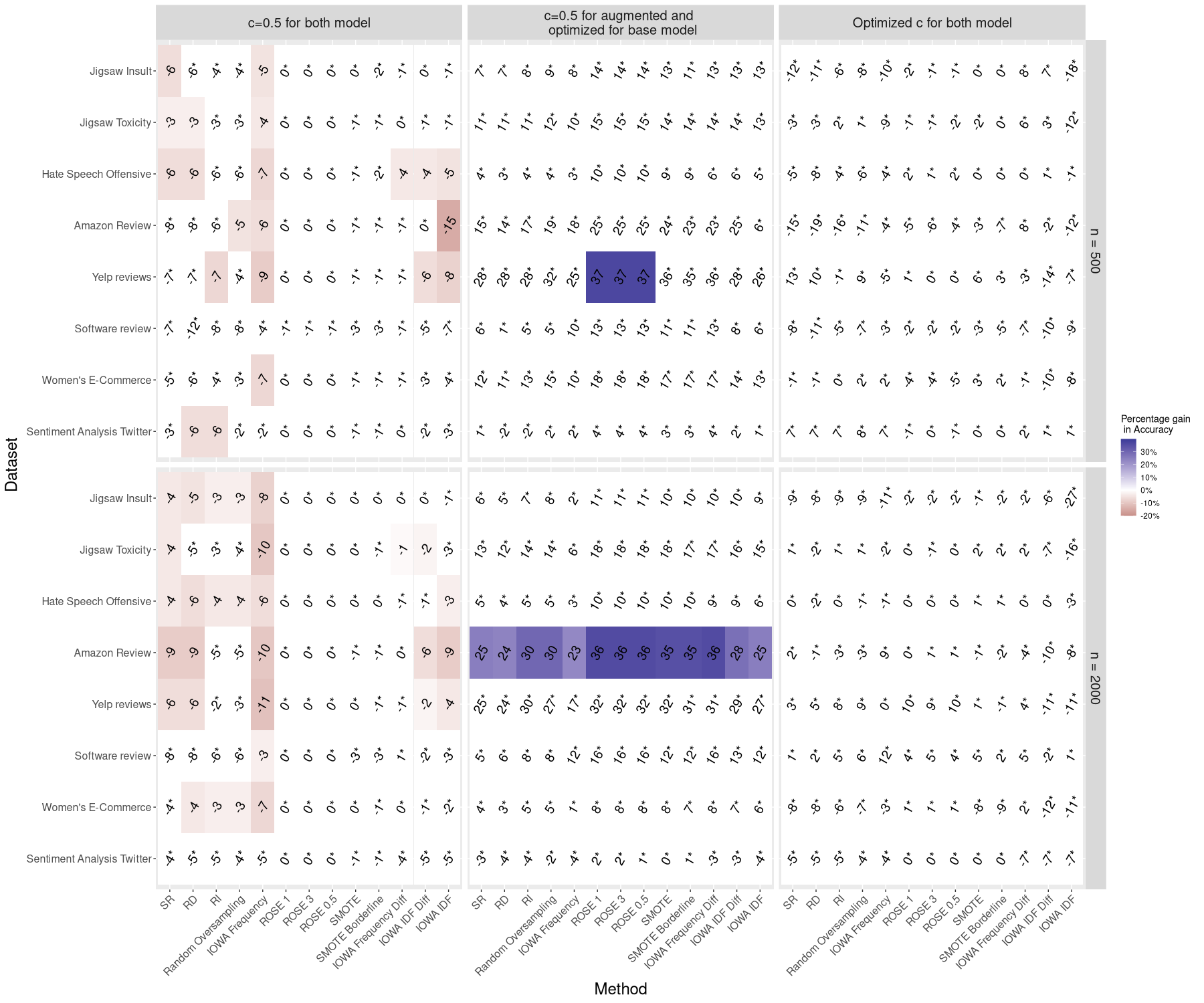}
    \caption{Heatmap of the mean percentage gain in Accuracy when comparing the augmented methods with the non-augmented model for classification rules.  Positive values indicate superior performance by the augmented method. Non-significant gains are marked with asterisks and displayed in white. For this metric when using the default threshold the non-augmented method has a better result, and when optimizing the threshold in a few cases has an increase for the augmented methods.}
\end{figure}

\newpage
\bibliographystyle{apalike}
\bibliography{refs}